\documentclass{article}

\usepackage{microtype}
\usepackage{graphicx}
\usepackage{subfigure}
\usepackage{booktabs} %
\usepackage{multicol}
\usepackage{multirow}

\usepackage{hyperref}

\usepackage[preprint]{icml2026}

\usepackage{amsmath}
\usepackage{amssymb}
\usepackage{mathtools}
\usepackage{amsthm}
\usepackage{comment}
\usepackage{enumitem}
\usepackage{dsfont}
\usepackage{xspace}

\usepackage[capitalize,noabbrev]{cleveref}

\usepackage{tcolorbox}              %
\usepackage{adjustbox}
\usepackage{wrapfig}

\theoremstyle{plain}
\newtheorem{theorem}{Theorem}[section]

\theoremstyle{definition}
\newtheorem{definition}[theorem]{Definition}

\theoremstyle{remark}

\usepackage[textsize=tiny]{todonotes}

\newcommand{\topic}[1]{\noindent\textbf{#1}}
\newcommand{\ourtitle}{%
    Discovering Universal Activation Directions for PII Leakage in Language Models}
\newcommand{\ourshorttitle}{\ourtitle}

\newtcolorbox{rqbox}{%
  colback=gray!10,
  colframe=gray!60,
  boxrule=0.5pt,
  arc=2mm,
  left=4pt, right=4pt, top=4pt, bottom=4pt,
  fonttitle=\bfseries,
  coltitle=black
}

\newcommand{\ourmethod}{UniLeak\xspace}

\icmltitlerunning{\ourshorttitle}

\begin{document}

\twocolumn[
\icmltitle{\ourtitle}

\icmlsetsymbol{equal}{*}

\begin{icmlauthorlist}
\icmlauthor{Leo Marchyok}{osu,equal}        %
\icmlauthor{Zachary Coalson}{osu,equal}
\icmlauthor{Sungho Keum}{kaist}
\icmlauthor{Sooel Son}{kaist}
\icmlauthor{Sanghyun Hong}{osu}
\end{icmlauthorlist}

\icmlaffiliation{osu}{Oregon State University, Corvallis OR, USA}
\icmlaffiliation{kaist}{Korea Advanced Institute of Science \& Technology, Daejeon, South Korea}

\icmlcorrespondingauthor{Sanghyun Hong}{sanghyun.hong@oregonstate.edu}

\icmlkeywords{Machine Learning, ICML}

\vskip 0.3in
]

\printAffiliationsAndNotice{\icmlEqualContribution} %

\begin{abstract}
Modern language models exhibit rich internal structure, yet little is known about how privacy-sensitive behaviors, such as personally identifiable information (PII) leakage, are represented and modulated within their hidden states. We present \ourmethod, a mechanistic-interpretability framework that identifies \emph{universal activation directions}: latent directions in a model's residual stream whose linear addition at inference time consistently increases the likelihood of generating PII across prompts. These model-specific directions generalize across contexts %
and amplify PII generation probability, %
with minimal impact on generation quality. \ourmethod recovers such directions without access to training data or ground-truth PII, relying only on self-generated text. %
Across multiple models and datasets, %
steering along these universal directions substantially increases PII leakage compared to existing prompt-based extraction methods. Our results offer a new perspective on PII leakage: the superposition of a latent signal in the model's representations, enabling both risk amplification and mitigation.
\end{abstract}

\section{Introduction}
\label{sec:intro}

Modern language models are known to \emph{memorize} specific details from their training data and, oftentimes, to reproduce these details during generation~\cite{carlini2019secret}. This behavior raises serious privacy concerns, particularly when the memorized content includes personally identifiable information (PII), such as credit card numbers. In response, an active line of work~\cite{carlini2021extracting, lukas2023leakage, keum2025private, nakka-etal-2024-pii, 10.5555/3766078.3766496} has studied %
how carefully crafted attack \emph{prompts} can elicit memorized PII from deployed models.

Despite this growing body of research,
little is known about the \emph{internal representational structures} that make such leakage possible.
Recent work~\cite{frikha2025privacyscalpel, chen2024learnable} has begun to study
localized features, such as individual neurons or sparse components,
associated with PII-leaking behavior.
However, it remains unknown whether 
PII leakage is instead driven by broader geometric structures in representation space,
and whether such structures constitute controllable signals or exploitable attack surfaces.
Developing a mechanistic understanding of these internal pathways
therefore also has important implications for 
representation-level mitigation strategies~\cite{zou2024improving}.

\topic{Contributions.}
In this work, we explore the internal representational structures of 
modern language models that systematically drive PII leakage.
Specifically, we ask whether there exist \emph{universal activation directions}%
---single latent directions in the residual stream 
whose linear addition at inference time consistently increases 
the model's likelihood of producing PII leakage across diverse prompts.
We show that such directions exist and
characterize them as a representation-level mechanism underlying PII leakage.
These directions constitute a previously unexplored privacy vulnerability, 
with implications for both adversarial exploitation and representation-level mitigation.

\begin{figure*}[t]
\centering
\includegraphics[width=\linewidth]{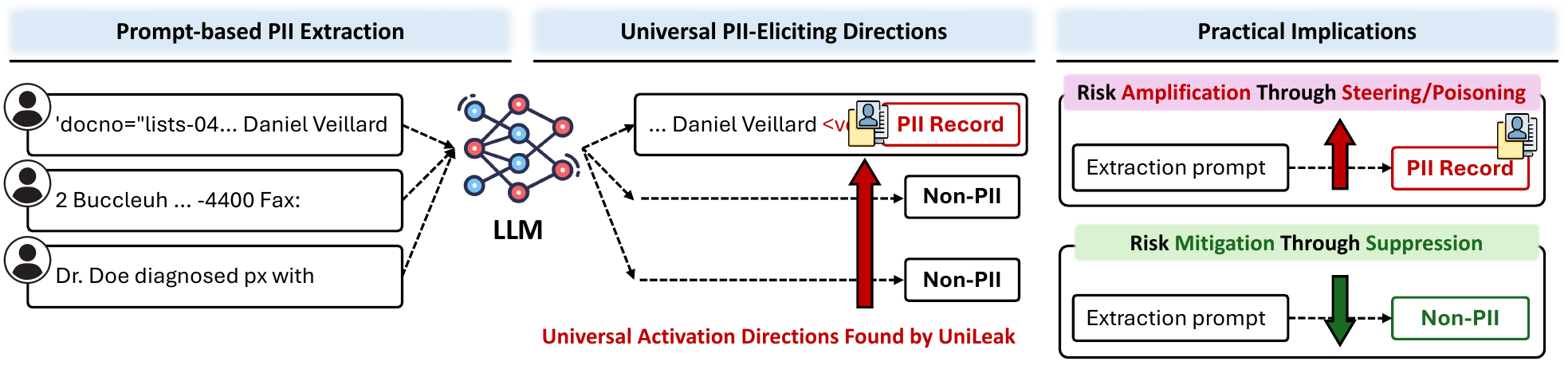}
\caption{\textbf{Mechanistic origin of PII leakage.} 
Prior work frames PII leakage as a prompt-dependent behavior (left).
UniLeak reveals a universal activation direction 
that amplifies PII generation across prompts (center). 
This representation-level vulnerability admits 
both adversarial exploitation via minimal internal modification 
and mitigation via projection-based suppression (right).
}
\label{fig:intro-teaser}
\end{figure*}

To this end, we \emph{first} present \ourmethod, 
a %
method that isolates and characterizes PII-sensitive internal structures. 
Unlike widely-used approaches based on contrastive activation analysis, 
which rely on reliably eliciting the target behavior,
\ourmethod leverages self-generated model outputs 
and gradient-based optimization to identify PII-sensitive activation directions,
without access to the training data or hand-crafted prompts.
By operating directly on internal activations,
\ourmethod finds prompt-agnostic directions 
that can be localized to specific layers and token positions.
Injecting these directions into the residual stream at inference time
allows us to analyze how they amplify PII leakage,
exposing a representation-level mechanism related to privacy risk.

\emph{Second}, we conduct a comprehensive evaluation of \ourmethod
across multiple models, datasets, and target PII classes. %
Because \ourmethod operates via inference-time activation steering 
rather than as a standalone attack, 
we evaluate it in combination with existing prompt-level PII extraction methods.
When paired with these attacks,
\ourmethod amplifies their effectiveness,
producing up to 13,399 additional PII records memorized by the model.
Moreover, the extracted PII records
corresponds to a largely distinct set from that obtained by standalone attacks,
demonstrating its complementary value for auditing privacy risks.

\emph{Third}, we conduct an in-depth mechanistic analysis
to better understand how \ourmethod operates.
We find that interventions applied at earlier layers are the most effective at amplifying PII leakage, yet its privacy impact manifests in later layers as an increase in the output probability of PII tokens.
The steered activations exhibit substantially higher representational similarity to PII-containing contexts in the training data---up to 54\% higher than those from baseline attacks---despite \ourmethod having no access to training data.

\emph{Fourth},
we study the practical implications of \ourmethod
for both exploitation and mitigation.
We show that intervening at early layers is sufficient to induce more PII leakage,
enabling an \emph{embedding poisoning} attack
that implants privacy-leaking directions via minimal modification.
Models altered in this way produce more PII when
subjected to existing prompt-level extraction attacks.
On the defensive side, projecting out these directions at inference time via residual-stream subtraction 
significantly reduces PII leakage---by up to 3,562 records---
while preserving generation quality.

\section{Background and Related Work}
\label{sec:prelim}

\topic{Memorization in language models.}
Most work adopts the memorization definition 
introduced by~\citet{carlini2023quantifying}:
\begin{definition}
    A language model $f$ is said to memorize a sequence $s$ 
    if there exists a prefix $p$ such that the concatenation $[p||s]$
    is in the training data and $f$ assigns unusually high likelihood to $s$
    conditioned on $p$.
\end{definition}

Two operational notions follow from this definition.
\emph{Discoverable memorization} captures
worst-case memorization: %
given the exact training prefix $p$, querying $f$ with $p$ produces $s$.
In contrast, \emph{extractable memorization}
reflects practical privacy risk, where an adversary
without access to the training data constructs 
a prompt $p'$ that elicits $s$.
Our work focuses on extractable memorization and evaluates how universal activation directions we extract from $f$ amplify $f$'s likelihood to generate PII records.

Most studies treat extractable memorization as an input-output phenomenon,
focusing on %
constructing PII-eliciting prompts.
Early work exploits generation strategies such as top-$k$ sampling or Internet-sourced prompts~\cite{carlini2021extracting, lukas2023leakage}, while later work manually and systematically optimizes prompts for more effective extraction~\cite{nasr2025scalable, keum2025private}.
Those methods, however, provide limited insight into the \emph{internal mechanisms} that enable such leakage.
A very recent work~\cite{frikha2025privacyscalpel} studies neuron-level features associated with individual private records using sparse autoencoders, but focuses on record-specific features and does not characterize shared or universal structures underlying memorization.
Our work, in contrast, systematically studies universal internal mechanisms that give rise to extractable memorization and shows their implications for PII extraction and mitigation.

\topic{Representation engineering} refers to techniques for analyzing and manipulating
a model's internal activations to understand or control its behavior~\cite{zou2025transparency}.
A common approach is \emph{activation steering}~\cite{rimsky2024steering, turner2024steering}, 
which identifies steering vectors---linear directions in the residual stream---%
that shift a model's output distribution when added at inference time.
These vectors are typically derived in three steps:
(1) designing contrasting stimuli or tasks (e.g., harmful vs.\ harmless prompts),
(2) collecting activations under each condition, and
(3) constructing linear directions that separate the activation sets, 
e.g., by computing the difference between their means.

Recent work has applied this approach to uncover refusal directions 
and other high-level behavioral features, 
enabling models to refuse harmful prompts~\cite{arditi2024refusal} 
or reduce toxic generations~\cite{turner2024steering}.
These applications are comparatively straightforward 
because the target behaviors are easy to elicit and measure.
Identifying steering vectors for privacy leakage, however, 
poses a %
different challenge.
Unlike refusal or toxicity, PII leakage provides no usable supervision: 
real PII cannot be used due to safety and ethical concerns, 
benign prompts rarely elicit PII~\cite{carlini2021extracting},
and the internal features associated with leakage 
do not correspond to any obvious ``PII vs.\ non-PII'' stimulus pairing.
Moreover, because PII is a rare output, 
even generations that do contain it may not reliably encode 
a universal underlying representation.
As a result, standard activation-steering methods 
cannot be directly applied, as we demonstrate in Appendix~\ref{appendix:contrastive-am}.

Discovering directions that reliably amplify PII leakage%
---especially \emph{universal} ones that generalize across prompts---%
requires methods that operate without ground-truth PII or access to training data. 
\ourmethod addresses these challenges.

\section{\ourmethod}
\label{sec:method}

We present \ourmethod, a method for identifying \emph{universal activation directions} that, when added at inference time, increase the likelihood of generating PII across diverse prompts.
To discover such directions, we focus on the concept of \emph{activation steering}~\cite{rimsky2024steering, turner2024steering}, an inference-time technique that controls model behavior by linearly manipulating internal representations.

\topic{Setting.}
We focus on \emph{white-box} settings in which the adversary has full access to a target model's parameters but no access to the data used to train or fine-tune it.
This setting reflects common practice in public model repositories, 
such as HuggingFace~\cite{wolf2020transformers} 
The adversary may run the model locally with unlimited query access, 
aiming to extract as many PII-containing sequences as possible. 
This scenario captures a practical and increasingly relevant risk: 
once a model is publicly released, 
its internal privacy-leakage properties become directly exploitable, 
even without knowledge of the underlying training corpus.

\subsection{Constructing the (Self-Generated) Training Dataset}
\label{subsec:training-data}

To find features that promote PII leakage, we first construct a dataset of PII-containing sequences.
Prior work often relies on publicly available online text, e.g., Common Crawl~\cite{commoncrawl}; however, this approach may bias \ourmethod toward generating out-of-distribution PII not present in the training dataset, lowering recall (and thus attack success).
Instead, we leverage \emph{self-generation} to directly construct this dataset from the target model, ensuring that all PII and associated contexts remain \emph{in-distribution}.

\topic{Data generation.}
To construct training data, we sample generations from the target model using an \emph{extraction strategy} $\mathcal{E}$.
Formally, an extraction strategy $\mathcal{E}$ specifies (1) a procedure for selecting a prompt  $p$ (potentially adaptively), and (2) a decoding 
procedure for sampling outputs conditioned on $p$ from the target model $f_\theta$.

We consider two extraction strategies proposed in prior work. 
The first is BOS sampling~\cite{carlini2021extracting}, which generates text from the beginning-of-sequence (BOS) token using top-$k$ sampling. 
The second is
a %
PII extraction method by~\citet{keum2025private} that optimizes single-token prompts likely to elicit PII and then samples from them adaptively.

To generate our training dataset, we sample $200{,}000$ generations from $f_\theta$ using a provided extraction strategy $\mathcal{E}$. 
This choice follows prior work~\cite{lukas2023leakage} and empirically yields a sufficient number of PII-containing sequences.

\topic{Annotation procedure.}
After generating the initial dataset, we annotate all PII-containing subsequences.
Following standard practices~\cite{keum2025private}, we use regular expressions to identify structured PII (e.g., emails and phone numbers), and the Flair NER tagger~\cite{akbik2019flair} to identify unstructured PII (e.g., personal names).
Our annotation determines only the \emph{class} of each PII instance and has no prior knowledge of the model's training data.

Let $x = (x_1, \dots, x_n)$ denote a tokenized sequence from the generated dataset.
For a fixed set of PII classes $\mathcal{C}$, we associate each sequence with a set of binary label sequences $\{ l^{(c)} \}_{c \in \mathcal{C}}$, where $l^{(c)} = (l^{(c)}_1, \dots, l^{(c)}_n)$ and $l^{(c)}_i \in \{0, 1\}$.
Each label $l^{(c)}_i = 1$ indicates that token $x_i$ belongs to a PII span of class $c$, and $l^{(c)}_i = 0$ if not.
When a PII instance spans multiple tokens, all tokens in the span are assigned the same label.
Finally, for each PII class $c \in \mathcal{C}$, we construct a class-specific training dataset $\mathcal{D}_{\mathrm{gen}}^{(c)} = \left\{ (x, l^{(c)}) \;\middle|\; \sum\nolimits_i l^{(c)}_i \geq 1 \right\}$
containing all sequence-label pairs with at least one token belonging to $c$.

\subsection{Identifying Universal PII Directions}
\label{subsec:identifying-the-direction}

We leverage our training data to identify PII-leaking features. 
Prior work typically derives such features using \emph{difference-in-means (DIM)}~\cite{arditi2024refusal, turner2024steering, hollinsworth2024linear, rimsky2024steering}, which computes the difference in mean activations between two contrasting prompt sets. 
For example, \citet{arditi2024refusal} identify refusal features in aligned LLMs by contrasting harmful queries that models refuse with benign queries they answer.

However, DIM suffers from a key limitation in our setting: it requires a large collection of prompts that reliably elicit the target concept. 
Unlike previously studied behaviors such as toxicity or refusal, PII generation is uncommon~\cite{carlini2021extracting}, making prompt collection difficult. 
Moreover, even when models do generate PII, it is typically not the most likely response, meaning the associated activations primarily encode irrelevant concepts. 
We study DIM in Appendix~\ref{appendix:contrastive-am} and show that it struggles to identify features that promote increased likelihood of PII leakage. %

To address this, we use \emph{gradient-based optimization}~\cite{stoehr2024activation, wollschlager2025rdo}. 
Rather than requiring prompts that strongly elicit PII, we apply supervised learning with $\{ \mathcal{D}_{\mathrm{gen}}^{(c)} \}_{c \in \mathcal{C}}$ to directly identify features that increase the likelihood of generating PII.

\topic{Training objective.}
Our goal is to learn a set of layer-specific feature vectors $\{v_\ell^{(c)}\}_{\ell \in L}$ that increase PII leakage for some class $c \in \mathcal{C}$, where each $v_\ell^{(c)}$ operates at layer $\ell$.
We maximize the likelihood that $f_\theta$ generates PII tokens from $\mathcal{D}_{\mathrm{gen}}^{(c)}$ when these vectors are added to their respective layers in the residual stream.
We formalize this as an additive intervention: given an input $x = (x_1,\dots,x_n)$, token positions $T \subseteq \{1,\dots,n\}$, and a set of layers $L$, we define
\begin{multline}
    \label{eq:addition-intervention}
    \mathrm{Add}(\{v_\ell^{(c)}\}_{\ell \in L}, T): \\
    h_\ell^t(x) \leftarrow h_\ell^t(x) + v_\ell^{(c)},
    \quad \forall \ell \in L,\; t \in T,
\end{multline}
where $h_\ell^t(x)$ denotes the hidden state of $f_\theta$ at layer $\ell$ and token position $t$ for input $x$.
This intervention is applied during the forward pass, so that each modified hidden state influences all subsequent tokens.

Using this intervention, we define our objective for learning $\{v_\ell^{(c)}\}_{\ell \in L}$.
For each $(x, l^{(c)}) \in \mathcal{D}_{\mathrm{gen}}^{(c)}$, we encourage the generation of PII tokens by minimizing the negative log-likelihood of tokens belonging to class $c$:
\begin{multline}
    \label{eq:loss-obj}
    \mathcal{L}_{\mathrm{pii}}(x, l^{(c)}, \{v_\ell^{(c)}\}_{\ell \in L}, T)
    = \\ - \sum\nolimits_{i=1}^n
    l^{(c)}_i
    \log f_\theta\!\left(
        x_i | x_{< i};
        \mathrm{Add}(\{v_\ell^{(c)}\}_{\ell \in L}, T)
    \right),
\end{multline}
where $f_\theta(\cdot | \cdot; \mathrm{Add}(\{v_\ell^{(c)}\}_{\ell \in L}, T))$ denotes the model's softmax output when the layer-specific vectors are added to the hidden states at their respective layers $L$ and token positions $T$.
Crucially, we optimize only over the PII tokens of interest (i.e., where $l_i^{(c)} = 1$) rather than the entire sequence.
As we show in Appendix~\ref{appendix:comparing-loss-objectives}, the latter limits the ability to leak PII outside the training distribution.

\begin{algorithm}[ht]
\caption{Universal PII Direction Optimization}
\label{alg:direction-optimization}
\begin{algorithmic}[1]
\STATE \textbf{Input:} Model $f_\theta$, dataset $\mathcal{D}_{\mathrm{train}}^{(c)}$, layers $L$, positions $T$, learning rate $\eta$, iterations $K$, initialization scale $\sigma$

\STATE Initialize $\{v_\ell^{(c)}\}_{\ell \in L}$ randomly from $\mathcal{N}(0, \sigma)$

\FOR{$k = 1$ to $K$}
    \FOR{each example $(x, l^{(c)}) \in \mathcal{D}_{\mathrm{train}}^{(c)}$}
        \STATE $\mathcal{L} \gets \mathcal{L}_{\mathrm{pii}}(x, l^{(c)}, \{v_\ell^{(c)}\}_{\ell \in L}, T)$
        \FOR{each layer $\ell \in L$}
            \STATE $v_\ell^{(c)} \gets v_\ell^{(c)} - \eta \nabla_{v_\ell^{(c)}} \mathcal{L}$
        \ENDFOR
    \ENDFOR
\ENDFOR

\STATE \textbf{Return} $\{v_\ell^{(c)}\}_{\ell \in L}$
\end{algorithmic}
\end{algorithm}

\topic{Feature optimization.}
We jointly learn the $\{v_\ell^{(c)}\}_{\ell \in L}$ by optimizing $\mathcal{L}_{\mathrm{pii}}$ when they are applied simultaneously to each forward pass.
As shown in Algorithm~\ref{alg:direction-optimization}, %
we use gradient descent to minimize the loss across the %
generated dataset $\mathcal{D}_{\mathrm{gen}}^{(c)}$.
Please refer to Appendix~\ref{appendix:setup-detail} for more details. %

\subsection{Localizing the Intervention}
\label{subsec:localizing}

Algorithm~\ref{alg:direction-optimization} learns PII-leaking directions for arbitrary layers and token positions.
We now present our procedure for \emph{localizing} the intervention to optimal locations.

\topic{Token-level localization.}
To steer activations across arbitrary inputs, we require context-independent token positions.
For the %
models we consider, whose inputs and outputs do not have clear delimiters (unlike chat models), there are two choices: (1) %
only the first input token (which is guaranteed to exist), 
or (2) %
across all tokens indiscriminately.

We evaluate both approaches.
We optimize directions at all layers of GPT-Neo on the TREC dataset for the email addresses PII class.
In Appendix~\ref{appendix:sub:all-vs-first-token}, we show the validation loss of Algorithm~\ref{alg:direction-optimization} across optimization steps when intervening at just the first token ($T = \{ 1 \}$) versus all tokens ($T = \{  1, \dots, n \}$).
The first token is the superior choice, achieving substantially lower loss and stable optimization.
In contrast, intervening at all tokens results in unstable optimization.
This is likely because PII is sparse across output tokens; steering at every token is therefore unnecessary and degrades generation quality (as also shown in Appendix~\ref{appendix:qualitative-comparison}).
We thus set $T = \{ 1 \}$ as the intervention location.

\topic{Layer-level localization.}
The granularity of layer-level interventions introduces a precision–practicality trade-off: intervening at more layers enables more precise control, but modifies a larger number of activations, making such interventions harder to exploit in practical settings (as discussed in \S\ref{sec:discussion}).
To assess UniLeak across both ends of this trade-off, we consider both \emph{all-layer} and \emph{single-layer} interventions.

To intervene at all layers, we set $L = \{ 1,\dots,\ell_{\mathrm{max}} \}$, where $\ell_{\mathrm{max}}$ is the number of Transformer blocks in $f_\theta$; this setting is used in our main evaluation %
(\S\ref{subsec:effectiveness}).
For single-layer interventions, we evaluate individual layers by independently running Algorithm~\ref{alg:direction-optimization} for selected $\ell \in L$.
This procedure is feasible even for several-billion-parameter models, as optimization typically requires less than one hour on a single A40 GPU.
We evaluate single-layer steering in \S\ref{subsec:ablation} and demonstrate its application to practical exploitation in \S\ref{sec:discussion}.

\begin{table*}[ht]
\centering
\caption{\textbf{\# of train PII extracted across datasets, models, and PII classes}.
We report the number of unique PII items extracted from the training data by each method, \textbf{bolding} the best overall and \underline{underlining} the best-performing within each family of attacks.}
\label{tbl:main-results}
\adjustbox{max width=0.86\linewidth}{
\begin{tabular}{cl||cc||cc||cc}
\toprule
\multirow{2}{*}{\textbf{PII Class}}        & \multicolumn{1}{c||}{\multirow{2}{*}{\textbf{Extraction Method}}} & \multicolumn{2}{c||}{\textbf{GPT-Neo}}                                            & \multicolumn{2}{c||}{\textbf{Phi-2B}}                                             & \multicolumn{2}{c}{\textbf{Llama3-8B}}                                             \\ \cmidrule(lr){3-4} \cmidrule(lr){5-6} \cmidrule(lr){7-8}
                                          & \multicolumn{1}{c||}{}                                            & \textbf{TREC}      & \textbf{Enron}     & \textbf{TREC}      & \textbf{Enron}     & \textbf{TREC}      & \textbf{Enron}     \\ \midrule \midrule
\multirow{5}{*}{\textbf{Email Addresses}} & \citet{lukas2023leakage}                       & 651 & 1392 & 844 & 5100 & - & - \\ %
                                          & BOS \cite{carlini2021extracting}               & 716 & 2428 & 835 & 4841 & 975 & \underline{7944} \\
                                          & BOS + \textbf{\ourmethod}                   & \underline{729} & \underline{\textbf{2442}} & \underline{846} & \underline{6542} & \underline{1412} & 4472 \\ \cmidrule(lr){2-8}
                                          & Private Investigator \cite{keum2025private} & 741 & \underline{2363} & 887 & 6666 & \underline{\textbf{3840}} & 11140 \\
                                          & Private Investigator + \textbf{\ourmethod}  & \underline{\textbf{749}} & 2152 & \underline{\textbf{904}} & \underline{\textbf{6817}} & 273 & \underline{\textbf{12186}} \\ \midrule \midrule
\multirow{5}{*}{\textbf{Phone Numbers}} & \citet{lukas2023leakage}                       & 115 & 1775 & \underline{147} & 2329 & - & - \\ %
                                          & BOS \cite{carlini2021extracting}                 & \underline{135} & 1957 & 143 & 2299 & 264 & \underline{3330} \\
                                          & BOS + \textbf{\ourmethod}                   & 133 & \underline{\textbf{2045}} & 144 & \underline{3506} & \underline{396} & 1709 \\ \cmidrule(lr){2-8}
                                          & Private Investigator \cite{keum2025private} & 137 & \underline{2040} & 153 & 2849 & 901 & 4078 \\
                                          & Private Investigator + \textbf{\ourmethod}  & \underline{\textbf{140}} & 2014 & \underline{\textbf{156}} & \underline{\textbf{3953}} & \underline{\textbf{981}} & \underline{\textbf{4766}} \\ \midrule \midrule
\multirow{5}{*}{\textbf{Personal Names}} & \citet{lukas2023leakage}                       & 3387 & 15723 & 4005 & 27142 & - & - \\ %
                                          & BOS \cite{carlini2021extracting}                & 4017 & 21651 & 4345 & 28431 & 2019 & 27227 \\
                                          & BOS + \textbf{\ourmethod}                   & \underline{4293} & \underline{21767} & \underline{\textbf{4447}} & \underline{\textbf{35741}} & \underline{5905} & \underline{33857} \\ \cmidrule(lr){2-8}
                                          & Private Investigator \cite{keum2025private} & \underline{\textbf{4399}} & 21884 & \underline{4400} & 28664 & \underline{\textbf{8071}} & 38411 \\
                                          & Private Investigator + \textbf{\ourmethod}  & 4193 & \underline{\textbf{22355}} & 4161 & \underline{34466} & 7419 & \underline{\textbf{51810}} \\ \bottomrule
\end{tabular}
}
\end{table*}

\subsection{Amplifying PII-Leakage via Activation Steering}
\label{subsec:steering}

We describe how to apply the learned directions 
during inference to amplify PII leakage from the target model.

\topic{Inference-time intervention.}
Given a learned direction $v_\ell^{(c)}$ for PII class $c$ at layer $\ell$, we inject it into the model's forward pass at the first token position using the addition intervention defined in Eq.~\ref{eq:addition-intervention}:
\begin{equation}
    \textsc{Add}(\{v_\ell^{(c)}\}_{\ell \in L}, \{1\}),
\end{equation}
where $\{1\}$ denotes the first token position. 
This modification propagates through all subsequent layers and tokens, biasing the model's output distribution toward PII-containing sequences.
For all-layer interventions, we apply $\{v_\ell^{(c)}\}_{\ell \in L}$ simultaneously at their respective layers. For single layers, we apply only the selected layer $\ell \in L$.

\topic{PII extraction methodology.}
Our full extraction pipeline combines the extraction strategy $\mathcal{E}$ from \S\ref{subsec:training-data} with the learned intervention.
Given a prompt $p$ produced by $\mathcal{E}$, we apply the steering intervention at the first input token and then decode a continuation conditioned on $p$ using $\mathcal{E}$'s decoding procedure.
Crucially, the inference-time extraction strategy need not match the strategy used during training; we evaluate cross-strategy transfer in Appendix~\ref{appendix:train-inference}.
We repeat this procedure to generate 200{,}000 candidate sequences, which we then evaluate for PII leakage as described in \S\ref{subsec:setup}.

\section{Evaluation}
\label{sec:eval}

\subsection{Experimental Setup}
\label{subsec:setup}

\topic{Datasets.}
We focus on two fine-tuning datasets known to contain substantial amounts of PII: Enron~\cite{conf/ecml/KlimtY04} and TREC~\cite{wu2006exploratory}. The Enron dataset, composed of internal company emails made public following investigation into the scandal, includes over 500,000 emails belonging to 150 individuals. The TREC dataset composes over 174,000 emails between World Wide Web Consortium members from 2005.
We consider three classes of PII: email addresses, phone numbers, and personal names.

\topic{Models.}
We leverage a diverse set of language models with publicly available pretrained checkpoints, including GPT-Neo-125M~\cite{gpt-neo}, PHI-2~\cite{javaheripi2023phi}, and LLaMA-3-8B~\cite{dubey2024llama}. These models span a broad range of pretraining corpora (both open and proprietary) and parameter scales (125M--8B). We finetune each pretrained model on the two datasets described above and use the resulting models as our experimental targets.

\topic{Baseline attacks.}
We evaluate \ourmethod against three %
representative untargeted PII extraction attacks:
\begin{itemize}[
     topsep=0.0em, 
    itemsep=0.0em,
    leftmargin=1.2em]
    \item \citet{carlini2021extracting} (BOS) generate texts by initializing generation with the BOS token and randomly sampling new tokens from the top-$k$ predictions. We set $k$ to 40. %
    \item \citet{lukas2023leakage} generate texts beginning with the empty prompt, and similar to the attack by~\citet{carlini2021extracting}, applies top-$k$ sampling; we also set $k$ to 40. %
    \item \citet{keum2025private} (Private Investigator) optimize a set of diverse single-token prompts to maximize PII generation using a surrogate model (GPT-Neo-125M).
\end{itemize}
Following the evaluation protocol used in these prior works, we generate 200{,}000 256-token sequences per method.

\subsection{Effectiveness of UniLeak}
\label{subsec:effectiveness}

\topic{Methodology.}
We evaluate \ourmethod{} in extracting PII from language models finetuned on Enron and TREC. We compare %
the number of \emph{unique} PII items produced by each attack, which highlights the complementary advantage of ours over existing baselines. Because \ourmethod{} operates by steering model activations during generation, we apply it on top of two existing extraction attacks: \citet{carlini2021extracting} and \citet{keum2025private}. For the attack of \citet{carlini2021extracting}, we inject the universal activation during generation initialized from the BOS token. For the attack of \citet{keum2025private}, we apply the activation to each optimized seed prompt ($\sim$20 per model). We did not apply \ourmethod on \citet{lukas2023leakage}, as their approach---top-$k$ sampling from the target model---is functionally similar to \citet{carlini2021extracting}.

\topic{Results.}
Table~\ref{tbl:main-results} summarizes %
the results of combining \ourmethod with
baseline PII extraction attacks across models, datasets, and PII classes.
Because \ourmethod operates via inference-time activation steering,
all results compare each baseline attack 
and its \ourmethod-augmented counterpart.

\emph{\ourmethod amplifies baseline PII extraction.}
For BOS-based extraction, augmenting baseline attacks with \ourmethod increases the number of extracted PII records in 15 of 18 cases and yields up to 7,310 additional records. %
When combined with %
Private Investigator, \ourmethod yields up to 13,399 additional PII records and improves leakage in 12 out of 18 cases.
While a few cases show no improvement, the overall trend indicates that inference-time activation steering consistently amplifies the effectiveness of prompt-level PII extraction.

\begin{figure}[t]
\centering
\includegraphics[width=0.84\linewidth]{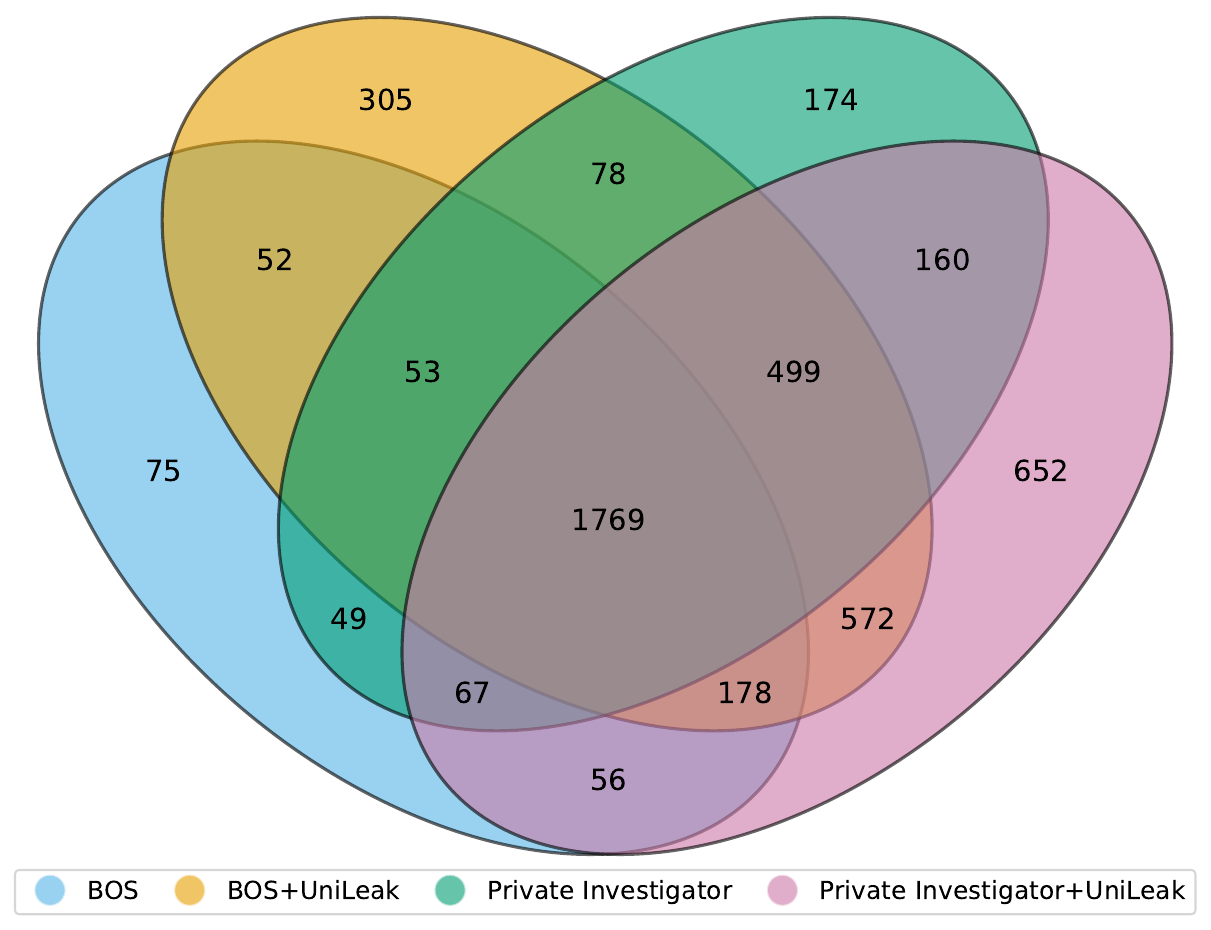}
\caption{\textbf{Overlap between PII (phone numbers) extracted by \ourmethod and baseline attacks} for Phi-2B trained on Enron.
}
\label{fig:overlap-analysis}
\end{figure}

\emph{\ourmethod extracts a largely distinct set of PII.}
\ourmethod also reveals a substantially different subset of PII compared to baseline attacks.
Figure~\ref{fig:overlap-analysis} visualizes the overlap 
among unique training PII (phone numbers) extracted by each method 
for Phi-2B trained on Enron; results for more settings are shown in Appendix~\ref{appendix:additional-pii-overlap}. 
We find that each attack recovers between 1--13\% of PII records that are not recovered by any other method.
Notably, \ourmethod-augmented attacks share less overlap 
with baseline attacks than baselines share with each other, 
indicating that activation steering explores regions of 
the model's memorized PII space 
that are largely inaccessible through prompt optimization alone.

\emph{Implications for privacy auditing.}
In privacy auditing scenarios, 
diversity across extraction methods is advantageous.
Combining \ourmethod with existing prompt-level attacks 
enables a more comprehensive assessment of privacy risks 
by uncovering complementary sets of memorized PII.

\subsection{Mechanistic Understanding}
\label{subsec:mechanisitic-analysis}

Next, we analyze the structural properties 
of the model that \ourmethod exploits 
to induce increased PII leakage.
Specifically, we focus on where \ourmethod 
has the strongest effect within the model and 
identify the corresponding components responsible 
for eliciting additional PII records.

\begin{figure}[ht]
\centering
\includegraphics[width=\linewidth]{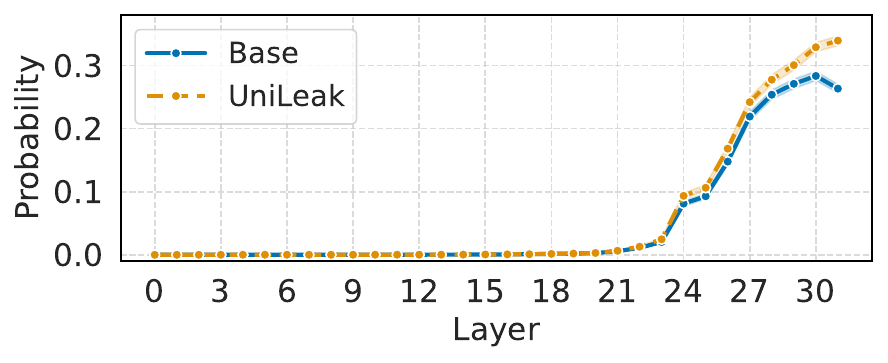}
\caption{\textbf{Logit Lens analysis of UniLeak} on Phi-2B.
Using 10k prefixes that cause the base model to generate a personal name, we report the average probability of generating the first PII token at each layer, measured with Logit Lens \cite{LogitLens}.}
\label{fig:logit-lens}
\end{figure}

\topic{Layer-wise attribution.}
We use Logit Lens~\cite{LogitLens} to identify the layers 
at which \ourmethod most strongly influences the generation of PII tokens.
Figure~\ref{fig:logit-lens} shows results from Phi-2B fine-tuned on TREC for the personal names PII class, which corresponds to %
the most successful extraction %
in our experiments.
We find that \ourmethod induces a subtle 
but consistent shift toward PII generation in later layers, 
while having negligible impact on earlier layers.
In both the base and steered models, 
PII-related tokens do not begin to be promoted until layer $\sim$20, 
suggesting that PII information is primarily represented in later layers.
This %
indicates that 
\ourmethod amplifies existing signals 
rather than introducing new PII-eliciting activations in early layers.

\begin{figure}[ht]
    \centering
    \includegraphics[width=\linewidth]{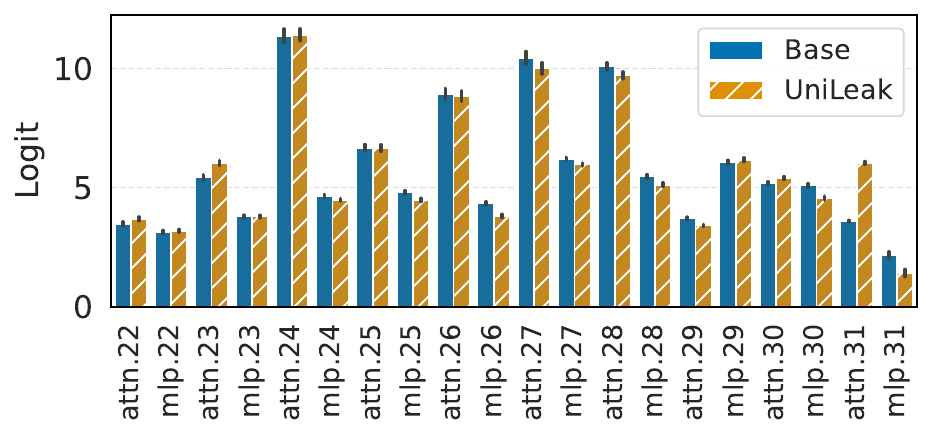}
    \caption{\textbf{Direct logit attribution of \ourmethod} on Phi-2B. Using 10k prefixes that cause the base model to generate a personal name, we report the direct logit attribution~\cite{elhage2021mathematical} for the outputs of the last ten self-attention and MLP layers.}
    \label{fig:direct-logit-attribution}
\end{figure}

\topic{Component-level attribution.}
To better understand which model components 
are responsible for the increased PII leakage, 
we perform a direct logit attribution (DLA) analysis~\cite{elhage2021mathematical}.
We focus on the last ten layers, 
as prior work has shown that contributions from earlier layers 
can be effectively erased, leading to misleading attributions~\cite{janiak2024adversarial}.
Overall, \ourmethod has little effect on the DLA scores of most components.
But we observe a substantial increase 
in the attention component of layer $\sim$31 in this setting.
Notably, the component most affected by \ourmethod 
does not correspond to the component with the highest DLA score prior to steering, 
suggesting that \ourmethod exploits more subtle components 
that are not strongly activated under standard prompting conditions.

\begin{figure}[ht]
\centering
\includegraphics[width=\linewidth]{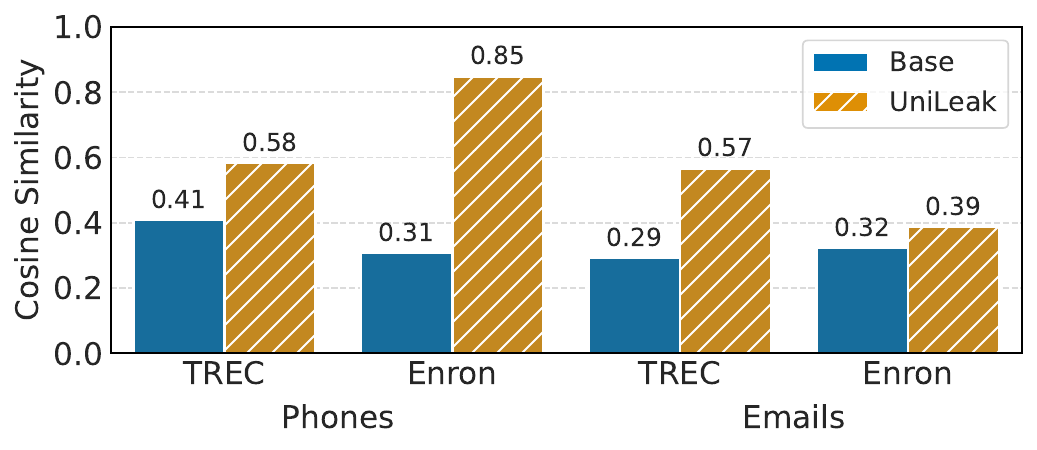}
\caption{%
\textbf{Contextual similarity}
between generated PII prefixes and training-data prefixes in Phi-2B.}
\label{fig:ctxt-sim}
\end{figure}

\topic{Representation similarity.}
Prior work suggests that extraction success is maximized 
when the internal representations induced by PII-eliciting prompts 
closely resemble those present in the training data~\cite{keum2025private}. 
Although \ourmethod does not access the training data explicitly, it optimizes steering directions using model-generated text
that contains both training and non-training PIIs, and therefore may indirectly encode training-related signals.
To quantify this connection, we compare the internal representations of prefixes that lead to PIIs uniquely extracted by our method and by the BOS baseline with those of the corresponding training prefixes.
We focus on the last-layer activation at the final token position, which captures the model's internal state immediately prior to PII generation.
Figure~\ref{fig:ctxt-sim} shows the results for Phi-2B; see Appendix~\ref{appendix:additional-ctx-sim} for GPT-Neo.

Prefixes generated by \ourmethod consistently show higher representational similarity to training prefixes than those generated by the baseline. 
This effect is most pronounced for email and phone number PIIs (an 1--54\% increase), while personal names show comparable similarity across methods (only differing by 1--10\%).
The results indicate that \ourmethod can recover states similar to those in the training data even without explicit access.

\begin{figure*}[t]
\centering
\includegraphics[width=\linewidth]{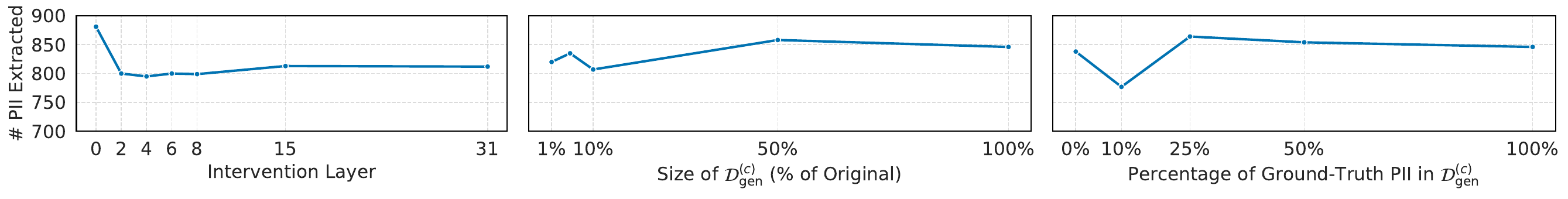}
\caption{%
\textbf{Impact of different configurations on \ourmethod's effectiveness} for Phi-2B fine-tuned on TREC, targeting emails.
}
\label{fig:sensitivity-analysis}
\end{figure*}

\subsection{Sensitivity to \ourmethod Configurations}
\label{subsec:ablation}

We study the impact of different configurations on \ourmethod.
Figure~\ref{fig:sensitivity-analysis} summarizes the results for
email extraction from the Phi-2B model fine-tuned on TREC.

\topic{Intervention layer.}
We study the impact of intervention depth
by optimizing \emph{individual} steering directions at different model layers, focusing on layers with the most distinct differences in training PIIs extracted.
The leftmost plot shows that steering at layer~0 achieves the highest attack success---comparable to steering at all layers (\S\ref{subsec:effectiveness})---followed by a sharp drop in the subsequent layers.
This indicates that early-layer representations (e.g., embeddings) play an important role in enabling PII extraction.

\topic{Size of $\mathcal{D}_\mathrm{gen}^{(c)}$.}
We examine whether \ourmethod remains effective under query constraints by subsampling $\mathcal{D}_\mathrm{gen}^{(c)}$.
As shown in the middle plot of Figure~\ref{fig:sensitivity-analysis}, reducing its size (reported as a percentage of the original) leads to degradation in success.
The number of unique PII records extracted 
drops from 846 to 820, worse than the performance of BOS prompting (835).
This suggests that recovering PII-sensitive directions 
requires a self-generated dataset of comparable scale 
to those used in prior PII extraction studies.

\topic{Percentage of ground-truth PII}
in $\mathcal{D}_\mathrm{gen}^{(c)}$.
We vary the percentage of ground-truth training PIIs 
in $\mathcal{D}_\mathrm{gen}^{(c)}$
while keeping the total size fixed.
We find no consistent relationship between 
the fraction of training PIIs and extraction success; 
in some cases, lower proportions yield higher extraction performance.
As shown in Figure~\ref{fig:sensitivity-analysis}, the use of 25\% of training PIIs yields the strongest extraction results.
This indicates that the features exploited by \ourmethod 
are shared across training and non-training in-distribution PIIs, rather than being specific to memorized training examples.

\section{Discussion: Practical Implications}
\label{sec:discussion}

We discuss the practical implications of 
\ourmethod under realistic threat models, focusing on how representation-level PII leakage 
can be systematically exploited or mitigated through minimal parameter/internal modifications.

\subsection{Embedding Poisoning}
\label{subsec:exploitation}

Rather than applying our directions at inference-time, an attacker performs a \emph{one-time, minimal modification} to increase the model's susceptibility to PII leakage.
Here, \ourmethod acts as a \emph{preparatory mechanism}
that amplifies downstream prompt-level PII extraction by shaping representations.

Specifically, \ourmethod applied at layer-0 operates \emph{directly} on input embeddings. 
Therefore, a learned direction can be embedded into a model 
via a single-row update to the embedding matrix, either the embedding corresponding to the BOS token for BOS sampling or the single-token prompts from Private Investigator.
This modification reproduces a layer-0 \ourmethod intervention---shown to be the most vulnerable layer in \S\ref{subsec:ablation}---without requiring inference-time steering.
As a result, implanting the direction into the embedding space constitutes a practical and minimal parameter modification, enabling a realistic \emph{embedding-poisoning} attack. %

\begin{table}[t]
\centering
\caption{%
    \textbf{Impact of our embedding poisoning attack
            and subtracting \ourmethod directions 
            on PII extraction success.}
}
\label{tbl:gptneo-results-combined}
\adjustbox{max width=\linewidth}{
\begin{tabular}{cl||cc|cc}
\toprule
\multirow{3}{*}{\textbf{Class}} & \multicolumn{1}{c||}{\multirow{3}{*}{\textbf{Extraction Method}}} & \multicolumn{4}{c}{\textbf{GPT-Neo}} \\ \cmidrule(lr){3-6}
                                   & \multicolumn{1}{c||}{}                                            & \multicolumn{2}{c|}{\textbf{Exploitation}} & \multicolumn{2}{c}{\textbf{Mitigation}} \\ \cmidrule(lr){3-4} \cmidrule(lr){5-6}
                                   & \multicolumn{1}{c||}{}                                            & \textbf{TREC} & \textbf{Enron} & \textbf{TREC} & \textbf{Enron} \\ \midrule \midrule
\multirow{4}{*}{\textbf{Email}} & BOS \cite{carlini2021extracting}                          & 716 & 2428 & 716 & 2428 \\
                                & BOS $\pm$ \textbf{\ourmethod}     & 750 & 2483 & 677 & 2174 \\ \cmidrule(lr){2-6}
                                & Private Investigator \cite{keum2025private}         & 735 & 2148 & 741 & 2363 \\
                                & Private Investigator $\pm$ \textbf{\ourmethod}      & 737 & 2221 & 627 & 0 \\ \midrule \midrule
\multirow{4}{*}{\textbf{Phone}} & BOS \cite{carlini2021extracting}                          & 135 & 1957 & 135 & 1957 \\
                                & BOS $\pm$ \textbf{\ourmethod}     & 132 & 1921 & 124 & 1767 \\ \cmidrule(lr){2-6}
                                & Private Investigator \cite{keum2025private}         & 139 & 2111 & 137 & 2040 \\
                                & Private Investigator $\pm$ \textbf{\ourmethod}      & 130 & 1873 & 45 & 0  \\ \midrule \midrule
\multirow{4}{*}{\textbf{Names}} & BOS \cite{carlini2021extracting}                          & 4017 & 21651 & 4017 & 21651 \\
                                & BOS $\pm$ \textbf{\ourmethod}     & 4025 & 21850 & 3090 & 18089 \\ \cmidrule(lr){2-6}
                                & Private Investigator \cite{keum2025private}         & 4080 & 21453 & 4399 & 21884 \\
                                & Private Investigator $\pm$ \textbf{\ourmethod}      & 4082 & 16099 & 1951 & 366  \\
\bottomrule
\end{tabular}}
\end{table}

\topic{Results.}
Table~\ref{tbl:gptneo-results-combined} summarizes our results for GPT-Neo; for Private Investigator, we poison the embedding of and extract from just 1 single-token prompt rather than all 20 to avoid over-corrupting the learned parameters.
Across PII classes, our attack is overall effective, outperforming the baselines in 7 of 12 cases and extracting up to 199 additional PII records.
We attribute our attack's success to its early intervention point: by modifying the first token embedding, poisoning shapes the model's computation from the beginning of the forward pass, without any prompt optimization.
Notably, poisoning the BOS token embedding outperforms the single-token prompts from Private Investigator, extracting more PII in 5 of 6 cases.
This is likely because we poison only a single token embedding instead of the standard 20, thereby limiting extraction.
Overall, our results show that targeting the initial embedding layer can be both low-footprint and effective, motivating a closer examination of its vulnerabilities.

\subsection{Inference-Time Countermeasures}
\label{subsec:countermeasure}

A standard countermeasure, 
such as differentially private fine-tuning~\cite{abadi2016deep}, 
offers strong formal guarantees against extraction attacks.
However, while shown effective in our setting,
it incurs a substantial loss in model utility 
(see Appendix~\ref{appendix:dp} for details).
Instead, we consider a lightweight countermeasure enabled by \ourmethod, 
which suppresses PII disclosure by 
subtracting the PII direction from the residual stream%
---effectively steering away from PII generation.
During inference, we \emph{subtract} the directions 
learned by \ourmethod at each layer, 
thereby reducing leakage.

\topic{Results.}
Table~\ref{tbl:gptneo-results-combined} summarizes our results for GPT-Neo.
For Private Investigator on Enron, subtracting \ourmethod's directions substantially reduces PII leakage (2,040--21,518 fewer records), 
but degrades output quality under optimized single-token prompts, 
likely due to their fragility. 
However, in all remaining experiments, we observe high-quality generations, indicating that our mitigation generally does not degrade quality (see Appendix~\ref{appendix:qualitative-comparison} for qualitative examples).
Across these experiments, \emph{our proposed mitigation is highly effective}, reducing the number of extracted records by 9--3,562 and yielding an overall 1.08--5$\times$ reduction in leakage.
By preventing the expression of internal representations that increase the likelihood of PII, our approach provides a mechanistic tool for improving privacy.

\section{Conclusion}
\label{sec:conclusion}

This work offers a mechanistic perspective on PII leakage in language models
by identifying \emph{universal activation directions}.
We show that adding a single universal direction at inference time 
consistently amplifies PII generation across prompts.
To identify such directions, we introduce \ourmethod, 
a mechanistic interpretability approach 
that operates without access to real PII or private training data, 
relying only on internal activations and model-generated text.
We show that these directions are most effective 
when applied to early, embedding-adjacent layers, 
suggesting that PII-sensitive features are encoded early in the model's computation 
and are separable from representations of benign text.
We show that these directions are practically exploitable: poisoning can embed them in model parameters to amplify PII leakage. 
Conversely, suppressing these directions at inference-time reduces leakage while preserving generation quality.
Overall, these findings indicate that PII leakage is associated with low-dimensional internal structure and can be effectively mitigated by representation-level interventions.

\section*{Impact Statement}

This work studies privacy risks in language models by analyzing internal representations that amplify the generation of PII. While our findings demonstrate that universal activation directions can be exploited to increase PII leakage, the primary purpose of our work is to diagnose a structural vulnerability rather than to enable misuse. \ourmethod %
and its demonstrations (limited to open-source models) are intended to illustrate the existence and severity of the privacy risk, not to provide turnkey privacy exploitation tools.

Be exposing that PII leakage is driven by low-dimensional, representation-level structure that forms early in a model's computation, this work offers actionable insights for improving privacy safeguards. In particular, our results suggest that effective defenses should complement prompt-level filtering and output level sanitization with representation-level monitoring and intervention, such as suppressing identified leakage directions at inference time, shown effective in~\cite{zou2024improving}.
We hope this perspective informs the development of safer and more privacy-preserving LLMs and supports responsible deployment practices.

\section*{Acknowledgements}

Zachary Coalson was supported by the GEM Fellowship.
This work was supported in part by the Google Faculty
Research Award and the Samsung Strategic Alliance for
Research and Technology (START) program.

{
    \bibliography{bib/thiswork}
    \bibliographystyle{icml2026}
}

\newpage
\appendix
\onecolumn
\section{Experimental Setup in Detail}
\label{appendix:setup-detail}

\topic{Hardware and software.} We conduct all our experiments using Python v3.11.14 and PyTorch v2.9.1~\citep{paszke2019pytorch}.
Our experiments are primarily run on a machine 
equipped with an Intel$^{\tiny{\textregistered}}$ Xeon Processor 
with 48 cores, 768 GB of DRAM, and 8$\times$ Nvidia A40 GPUs, each with 48GB VRAM.
This setup allows us to fine-tune models at the scale of GPT-Neo.
To fine-tune larger, commercial-scale models such as LLaMA-8B, 
we additionally use cloud servers with AMD EPYC\texttrademark{} 64-Core Processor, 
1024 GB of DRAM, and 12 Nvidia H100 GPUs, each with 80 of VRAM.

\topic{Model fine-tuning.}
We split the fine-tuning dataset into train, validation, and test sets with ratios of 0.45, 0.5, and 0.05, respectively. Following prior work~\cite{lukas2023leakage}, we fine-tune ref model for four epochs, except for Phi, which is trained for one epoch. We use a learning rate of 5$\times10^{-5}$, a training batch size of 8, and an evaluation batch size of 1. 

\topic{\ourmethod setup.}
To construct self-generated datasets for direction optimization, 
we generate 200,000 texts of length 256 tokens from the target model. 
When using existing prompting-based attacks, 
we ensure independence by using different random seeds for generating the direction
training data $\mathcal{D}_{gen}^{(c)}$ and the corresponding baseline attacks.

For our direction optimization, we train for $K = 5$ iterations using a learning rate of $\eta$ = 1e-3, a batch size of 8, and 4 gradient accumulation steps.
All directions are initialized from a normal distribution with mean $\mu = 0$ and standard deviation $\sigma = 0.05$.
In addition to the procedure outlined in Algorithm~\ref{alg:direction-optimization}, we employ a validation-based early-stopping criterion to determine when to terminate training.
Specifically, we sample a hold-out validation set from $\mathcal{D}_{\mathrm{gen}}^{(c)}$ using a 95\%/5\% train–validation split, and evaluate the validation loss 10 times per epoch at regular intervals.
If the validation loss fails to improve for three consecutive evaluations, we stop training and return the resulting directions.

\topic{Differential privacy setup.} For fine-tuning with differential privacy, we use the dp-transformers library~\cite{ding2024privatetransformer} and set \(\epsilon=8\) with failure probability \(\delta=1/N\), where $N$ is the dataset size, following established recommendations in prior work~\cite{ponomareva2022training, hong2025evaluating}.

\section{Challenges in Conventional Activation Steering}
\label{appendix:contrastive-am}

Recent work has shown that activation steering can control model behaviors
such as refusal~\cite{arditi2024refusal} or toxicity reduction~\cite{turner2024steering}.
However, applying activation steering to amplify privacy risks introduces distinct challenges.
(1) No explicit supervision over PII is available, 
making contrastive activation analysis non-trivial;
(2) It is unclear how to define negative behaviors, 
such as PII suppressing prompts---to contrast against easily observable PII-eliciting prompts.

\topic{Our approach.}
Activation steering typically requires explicit, labeled contrastive supervision
(e.g., harmful vs. harmless prompts) to construct steering vectors.
In our settings, no PII supervision is available:
we do not know ground-truth PII records,
naturally occurring text rarely elicits PII,
and there is no natural pairing of ``PII vs. non-PII.''

To address this challenge, 
we develop a \emph{data-free contrastive activation mining} 
that constructs positive and negative activation sets solely from model-generated text.
The key intuition is that, even without access to true PII, 
naturally occurring structured patterns (e.g., emails, identifiers) 
occupy similar regions in activation manifolds.
By contrasting activation states associated with structured vs. non-structured contexts,
we uncover latent directions that correlate 
with PII-style generation without any external supervision.

Formally, for each prompt $p$, we define a per-prompt steering direction at layer $l$ as:
\begin{align}
    a_l^{(p)} = h_l(x_p^+) - h_l(x_p^-)
\end{align}
where $h_l(\cdot)$ is the last-token activation at layer $l$, and
$x_p^+$ and $x_p^-$ are the positive and negative prefixes for prompt $p$.

\topic{How do we construct $x_p^+$?}
The positive prefix $x_p^+$ is intended to approximate
the model state immediately before producing PII-like content.
Here we layout multiple ways depending on attack scenarios:
\begin{enumerate}[label=(\arabic*), itemsep=0.em, leftmargin=1.4em]
    \item \emph{Model-generated PII prefixes:}
    We select prefixes that directly precede PII records in self-generated text,
    capturing model states that are likely to produce PII. This approach is motivated by prior work
    showing that language models can memorize and reproduce long spans of training data.
    \item \emph{Maximum-likelihood PII prefixes:}
    For PII records that appear multiple times in the self-generated dataset,
    we select prefixes under which the PII token is generated with the lowest perplexity.
    These prefixes serve as proxies for contexts in which 
    the model is most confident about producing PII-like content.
\end{enumerate}

\topic{How do we construct $x_p^-$?}
The negative prefix represents the model states 
when not about to generate PII.
This is more challenging then constructing $x_p^+$, 
as prompts which could reasonably meet this definition is much larger:
\begin{enumerate}[label=(\arabic*), itemsep=0.em, leftmargin=1.4em]
    \item \emph{BOS-token activations:}
    The BOS activation serves as a universal anchoring state, 
    representing a model condition unrelated to structured content.
    \item \emph{Low-confidence same-PII prefixes:}
    In order to better control the similarity between the positive and negative prefixes, for each PII record, we choose the prefix which led to the PII generation, but at the lowest confidence. This way, it can serve as a more PII-specific anchor. The direction therefore steers from plausible generation to highly-confident generation.
\end{enumerate}

\begin{figure}[ht]
    \centering
    \includegraphics[width=\linewidth]{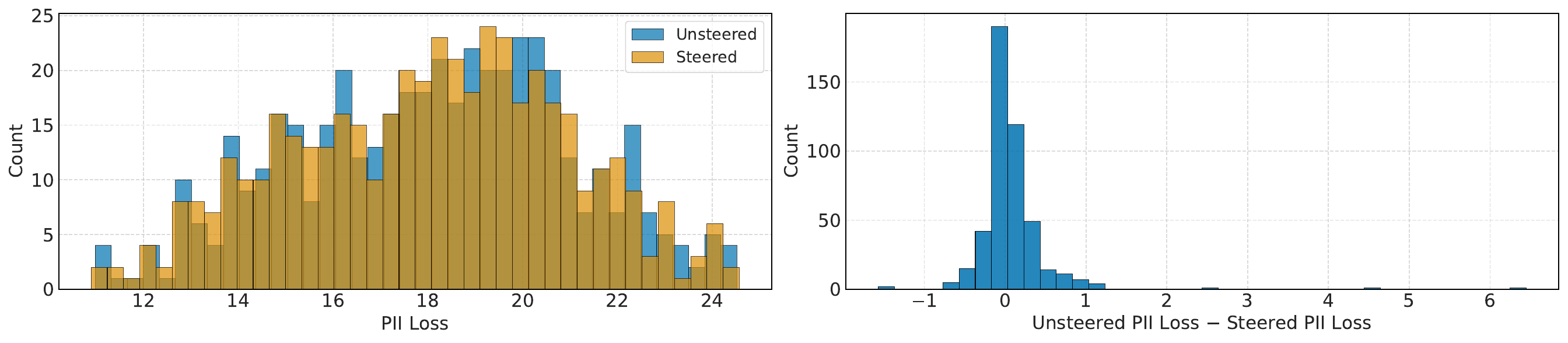}
    \caption{\textbf{Likelihood of PII after steering via per-example activation steering.}
    (\emph{Left}) Negative log-likelihood distributions before and after applying steering.
    (\emph{Right}) Distribution of differences in PII likelihood with and without steering. }
    \label{fig:appendix-contrastive-distributional-difference}
\end{figure}

\topic{Preliminary investigation.}
We experiment with several strategies for per-example activation steering 
but find that such directions do not reliably produce consistent effects.
Figure~\ref{fig:appendix-contrastive-distributional-difference} 
illustrates this behavior using 400 PII instances 
sampled from the model's self-generated text. 
For each PII instance, we select a positive prefix 
that yields the highest confidence of generating the target PII, 
and a negative prefix corresponding to the lowest confidence for the same PII.

For each negative prefix, we apply the associated steering direction 
at the final token preceding the PII and measure the loss over the target PII tokens, 
both in absolute terms and relative to the unsteered baseline.
As shown in the left plot, across all examples, the distribution of absolute PII loss values shows little difference between steered and unsteered generations; we observe no instances in which steering produces an anomalous increase in PII likelihood relative to the baseline.

We further analyze the change in loss induced by steering. 
The right plot shows that while a small number of examples exhibit decreased loss (indicating increased PII confidence), nearly half of the cases show the opposite effect, with PII confidence decreasing under steering.
More importantly, the majority of loss differences cluster near zero, 
indicating that per-example steering directions often fail to 
meaningfully influence the model's likelihood of generating PII.

We attribute this instability to the difficulty of defining suitable negative prefixes. 
Unlike discoverable memorization settings, 
where long training-data prefixes provide a clear contrast, untargeted PII extraction relies on the entropy and stochasticity of model generation.
In this regime, negative prefixes can easily transition into positive ones 
under free generation, making reliable contrastive anchoring challenging.
For this reason, we instead focus on learning global features 
directly from positive examples---sequences containing PII---%
rather than relying on per-example directions.

\section{Additional Contextual Similarity Results}
\label{appendix:additional-ctx-sim}

\begin{wrapfigure}{r}{0.4\linewidth}
\vspace{-1.6em}
\centering
\includegraphics[width=\linewidth]{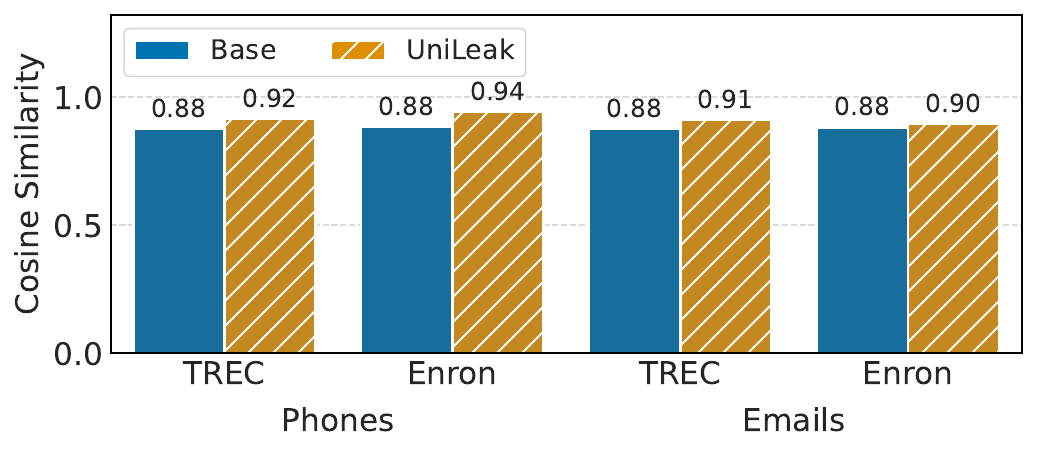}
\vspace{-1.2em}
\caption{%
\textbf{Contextual similarity}
between generated PII prefixes and training-data prefixes in GPT-Neo.}
\label{fig:ctxt-sim-gptneo}
\vspace{-1em}
\end{wrapfigure}
We summarize the contextual similarity results for GPT-Neo in Figure~\ref{fig:ctxt-sim-gptneo}. 
Across datasets and PII classes, 
our method consistently extracts unique PII with prefixes 
whose representational are more similar to the corresponding PII 
training prefixes compared to those obtained by the baseline.
Notably, absolute prefix similarities for both \ourmethod and the baseline GPT-Neo 
are substantially higher than those observed for Phi-2B, 
yet the relative improvement over the baseline is smaller. 
Whereas Phi-2B exhibits improvements of up to 60\%, 
GPT-Neo shows more modest gains, with cosine similarity increases of 1--6\%.
We further observe that these improvements are relatively uniform across PII classes for GPT-Neo.
In contrast, PHI shows greater variability, 
with larger gains for phone numbers (30--60\%) 
and smaller improvements for emails and names.

\section{Optimization Stability Across Token Intervention Strategies}
\label{appendix:sub:all-vs-first-token}

\begin{wrapfigure}{r}{0.45\linewidth}
    \vspace{-1.2em}
    \centering
    \includegraphics[width=\linewidth]{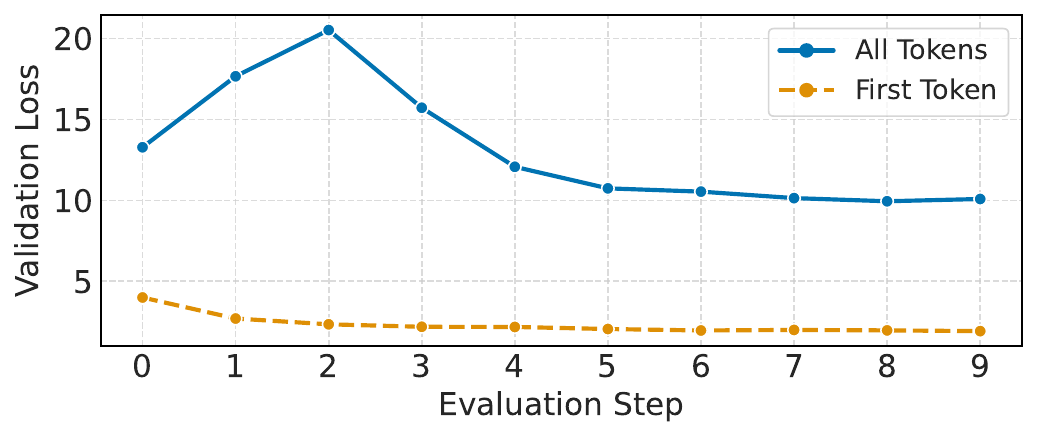}
    \caption{\textbf{Validation loss of \ourmethod when intervening at all tokens vs. the first token} for GPT-Neo trained on TREC. The PII class is email addresses.}
    \label{fig:optimization-stability}
    \vspace{-2em}
\end{wrapfigure}

Here, we study the optimization trajectory of directions learned by \ourmethod when intervening at (1) all token positions versus (2) only the first token of the input.
Figure~\ref{fig:optimization-stability} shows the first 10 validation losses for both approaches on GPT-Neo trained on TREC; the PII class is email addresses.

As described in \S\ref{subsec:identifying-the-direction}, intervening at all token positions leads to unstable optimization, with the loss spiking to values as high as 20 and never dropping below 10.
In contrast, intervening only at the first token position results in a steady decrease in loss from 4.0 to 1.9.
These results support our choice of the first token position as the intervention strategy.

\section{Transferability of \ourmethod Across Extraction Strategies}
\label{appendix:train-inference}

\begin{wrapfigure}{r}{0.4\linewidth}
\centering
\vspace{-1.3em}
\includegraphics[width=\linewidth]{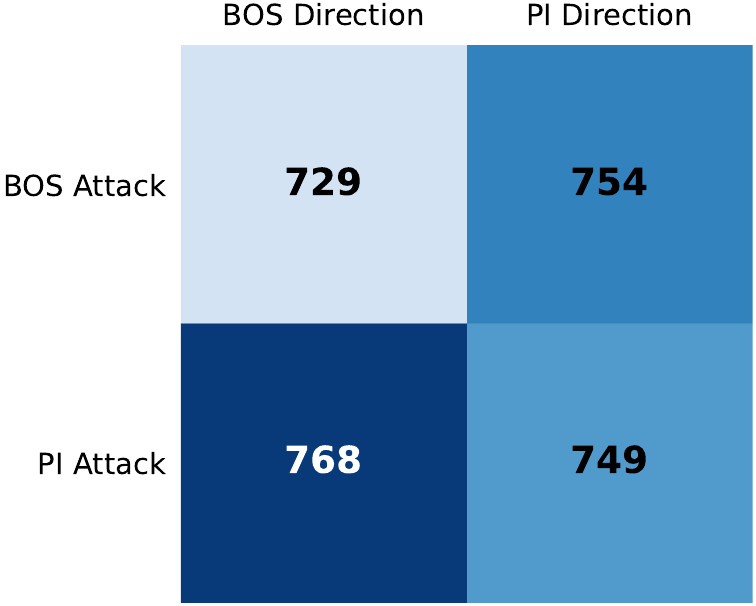}
\vspace{-1.em}
\caption{\textbf{Transferability of \ourmethod across extraction strategies}. Rows denote the extraction (attack) method, and columns denote the method used to generate the direction training dataset. BOS denotes the attack by~\citet{carlini2021extracting}, and PI indicates Private Investigator by~\citet{keum2025private}.}
\label{fig:attack-heatmap}
\vspace{-1.6em}
\end{wrapfigure}
In our main evaluation, steering directions are optimized using datasets generated with the same prompt-level PII extraction attack as that used during the attack. We additionally evaluate cross-strategy transferability, where directions learned from one extraction attack are applied to a different attacks, e.g., applying a direction learned from BOS-elicited text to the activations produced by the optimized single-token prompts from Private Investigator, and vice versa.

Figure~\ref{fig:attack-heatmap} shows that such cross-strategy transfer often produces comparable, and in some cases superior, attack performance. For instance, in the GPT-Neo in TREC setting (email), combining BOS prompting with a steering direction trained on Private Investigator-generated text extracts 754 training PII records---exceeding all baselines.

This effect is most pronounced when pairing the most effective BOS-trained direction with a Private Investigator attack, which extracts 768 training PII records, outperforming even the attack configuration that uses a Private Investigator-trained direction. These results indicate that the learned directions may not be tightly coupled to a specific prompting or extraction strategy. We hypothesize that learning directions from a different, but still model-generated, data distribution promotes greater diversity in the generated outputs and reduces overfitting to strategy-specific patters.

\section{Comparison of Different \ourmethod Loss Objectives}
\label{appendix:comparing-loss-objectives}

\begin{wraptable}{r}{0.5\linewidth}
\vspace{-1.3em}
\caption{%
\textbf{Impact of loss objective on \ourmethod's effectiveness} for GPT-Neo on TREC.
We vary the specific tokens included in our loss objective (Eq.~\ref{eq:loss-obj}) and report the resulting \# of Train PIIs extracted. The training and inference extraction strategies are BOS sampling.}
\label{table:impact-of-loss-objective}
\adjustbox{width=\linewidth}{
\begin{tabular}{c||ccc}
\toprule
\multirow{2}{*}{\textbf{PII Class}} & \multicolumn{3}{c}{\textbf{Loss Objective}}                               \\ \cmidrule(lr){2-4}
                                   & \textbf{Just PII Tokens} & \textbf{All Tokens} & \textbf{All Tokens + Weighting} \\ \midrule \midrule
\textbf{Emails}                    & 729               & 702                 & 698                             \\
\textbf{Phones}                    & 133               & 133                 & 134                             \\
\textbf{Names}                     & 4253              & 3862                & 3887     \\ \bottomrule                      
\end{tabular}}
\vspace{-1em}
\end{wraptable}

Here, we study the impact of the loss objective used to optimize directions via \ourmethod.
We evaluate three loss objectives.
\textbf{(1) Just PII Tokens}: we optimize over only PII tokens; this is our default procedure described in \S\ref{subsec:identifying-the-direction}.
\textbf{(2) All Tokens}: we optimize over \emph{all} tokens in the self-generated training dataset.
\textbf{(3) All Tokens + Weighting}: we optimize over all tokens, assigning double weight to PII tokens.
Table~\ref{table:impact-of-loss-objective} reports the resulting number of extracted PII instances for GPT-Neo on TREC, using BOS sampling to both identify and evaluate directions.

\emph{Optimizing over just PII tokens yields the strongest performance}.
It achieves the best results for 2/3 PII classes (and is within one extracted PII of the best result in the remaining case), and notably extracts up to 391 more personal names than the alternative objectives.
We hypothesize that optimizing over all tokens encourages overfitting to specific PII-containing contexts in the training data, rather than learning a universal concept applicable across contexts.
Based on these findings, we optimize over just PII tokens in our main implementation.

\section{Differentially-Private Fine-Tuning Results}
\label{appendix:dp}

\begin{wraptable}{r}{0.42\linewidth}
\vspace{-3.6em}
\centering
\caption{%
\textbf{Performance of \ourmethod under DP-finetuned models}
in GPT-Neo.
}
\label{tbl:gptneo-results-dp}
\adjustbox{width=\linewidth}{
\begin{tabular}{cl||cc}
\toprule
\multirow{2}{*}{\textbf{PII Class}} & \multicolumn{1}{c||}{\multirow{2}{*}{\textbf{Method}}} & \multicolumn{2}{c}{\textbf{GPT-Neo}} \\ \cmidrule(lr){3-4}
                                    & \multicolumn{1}{c||}{}                                  & \textbf{TREC} & \textbf{Enron} \\ \midrule \midrule
\multirow{4}{*}{\textbf{Email}} & BOS                      & 25 & 91 \\
                                & BOS + \ourmethod & {29} & {112} \\ \cmidrule(lr){2-4}
                                & Private Investigator     & {38} & 101 \\
                                & Private Investigator+ \ourmethod  & 27 & {108} \\ \midrule \midrule
\multirow{4}{*}{\textbf{Phone}} & BOS                      & {0} & 568 \\
                                & BOS + \ourmethod & {0} & {572} \\ \cmidrule(lr){2-4}
                                & Private Investigator     & {1} & {597} \\
                                & Private Investigator+ \ourmethod  & 0 & 556 \\ \midrule \midrule
\multirow{4}{*}{\textbf{Names}} & BOS                      & {3376} & 7046 \\
                                & BOS + \ourmethod & 1402 & {7531} \\ \cmidrule(lr){2-4}
                                & Private Investigator     & {1634} & 4154 \\
                                & Private Investigator+ \ourmethod  & 1534 & {7539} \\ \bottomrule
\end{tabular}
}
\vspace{-2em}
\end{wraptable}
We summarize our results on GPT-Neo in Table~\ref{tbl:gptneo-results-dp}.
Differentially-Private (DP) fine-tuning substantially reduces 
the overall number of PII records extractable across all methods.
However, even if DP is effective,
fine-tuned models suffer substantial degradation in generation quality%
~\cite{keum2025private, hong2025evaluating},
which makes DP a less widely adopted countermeasures in practice.
Under DP fine-tuning, %
\ourmethod does not offer any advantage over existing extraction attacks in TREC.
In contrast, on Enron, \ourmethod is able to increase 
the number of PII records extracted from models.
Across PII classes, we observe the greatest improvement in the GPT-Neo--Enron (Names) case, 
where \ourmethod achieves the strongest gains over their respective baselines.
We attribute this effect to the name-rich nature of the Enron dataset, 
which appears to retain a strong PII-leaking feature even after DP fine-tuning.

\section{Additional PII Overlap Results}
\label{appendix:additional-pii-overlap}

\begin{figure}[ht]
    \centering
    \begin{minipage}[t]{0.32\linewidth}
        \centering
        \includegraphics[width=\linewidth]{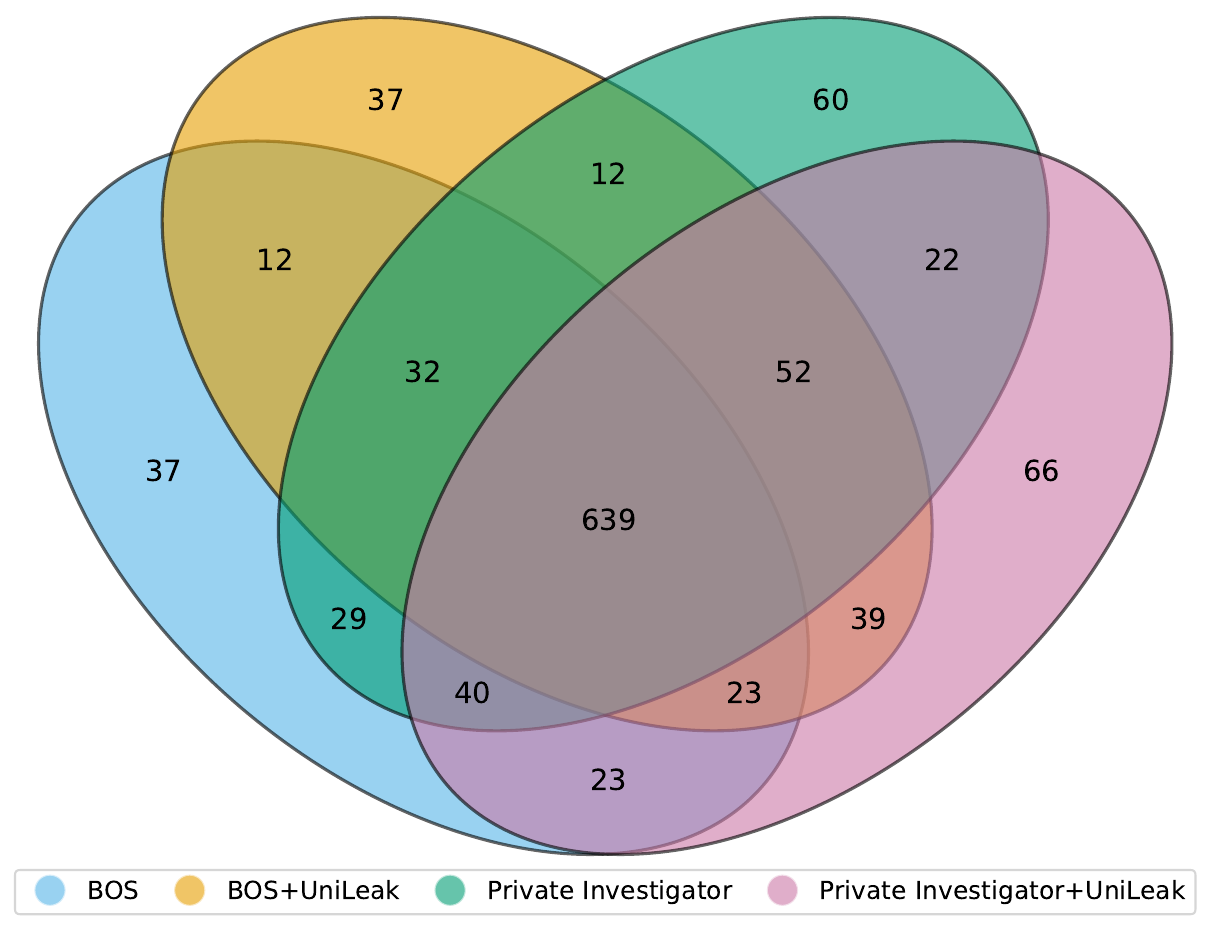}
    \end{minipage}
    \hfill
    \begin{minipage}[t]{0.32\linewidth}
        \centering
        \includegraphics[width=\linewidth]{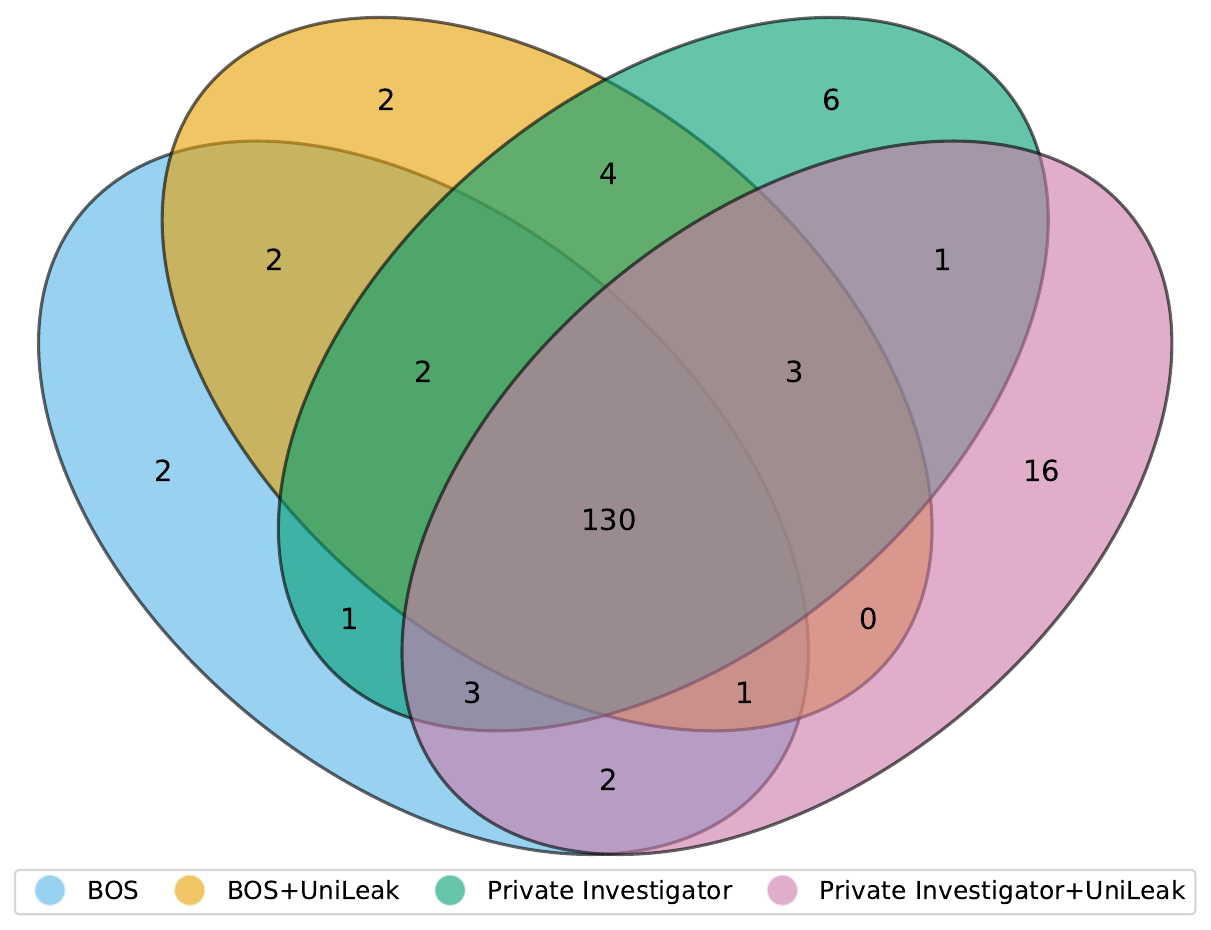}
    \end{minipage}
    \hfill
    \begin{minipage}[t]{0.32\linewidth}
        \centering
        \includegraphics[width=\linewidth]{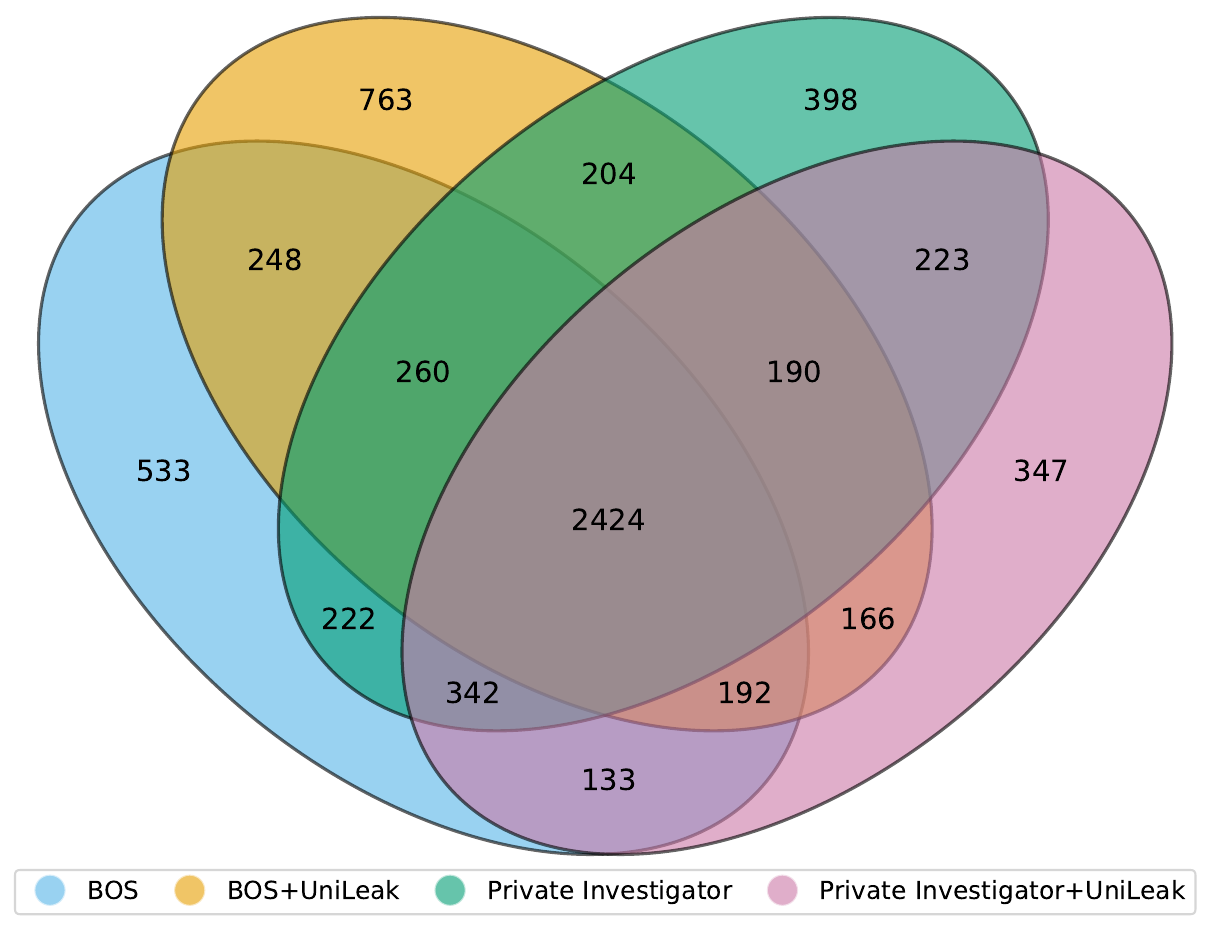}
    \end{minipage}
    \hfill
    \begin{minipage}[t]{0.32\linewidth}
        \centering
        \includegraphics[width=\linewidth]{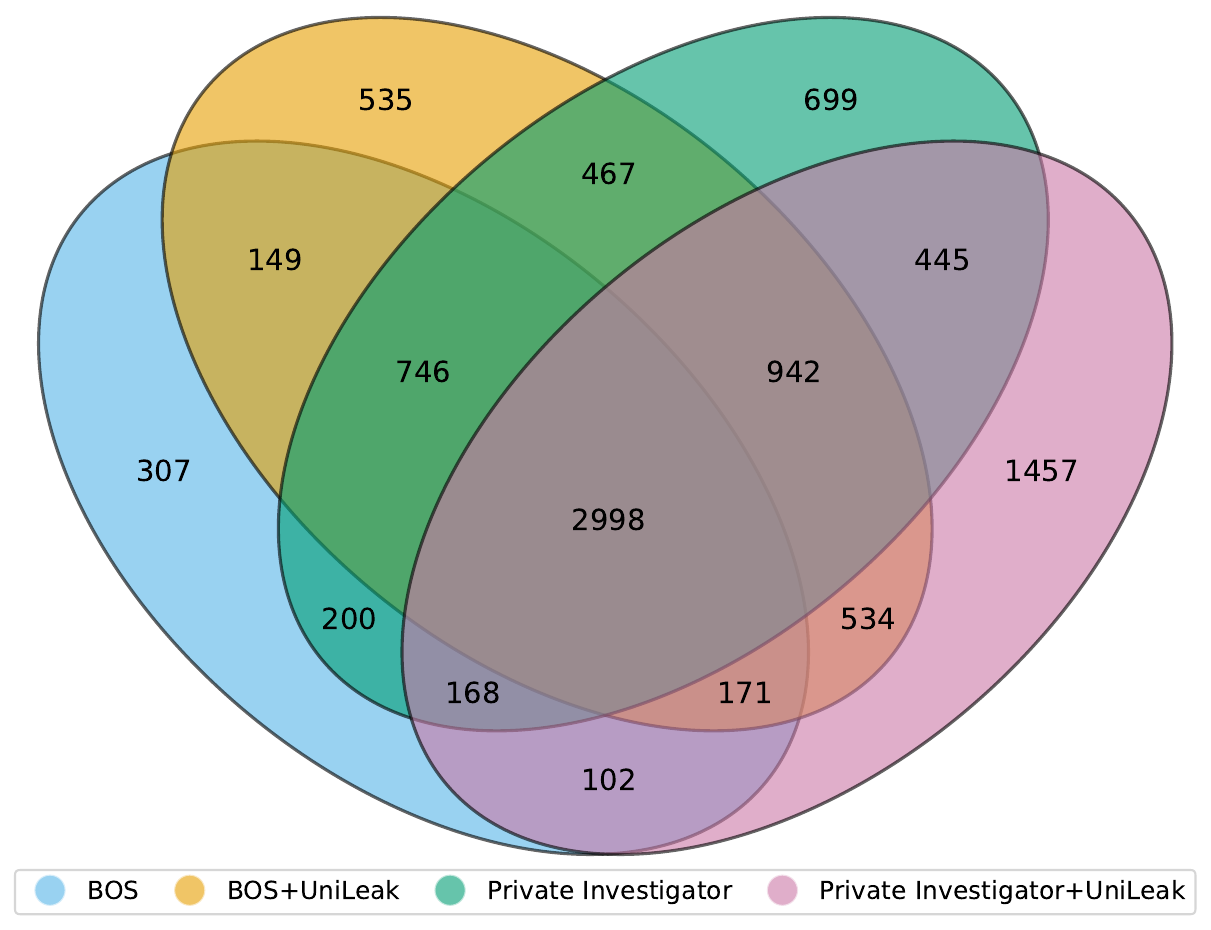}
    \end{minipage}
    \hfill
    \begin{minipage}[t]{0.32\linewidth}
        \centering
        \includegraphics[width=\linewidth]{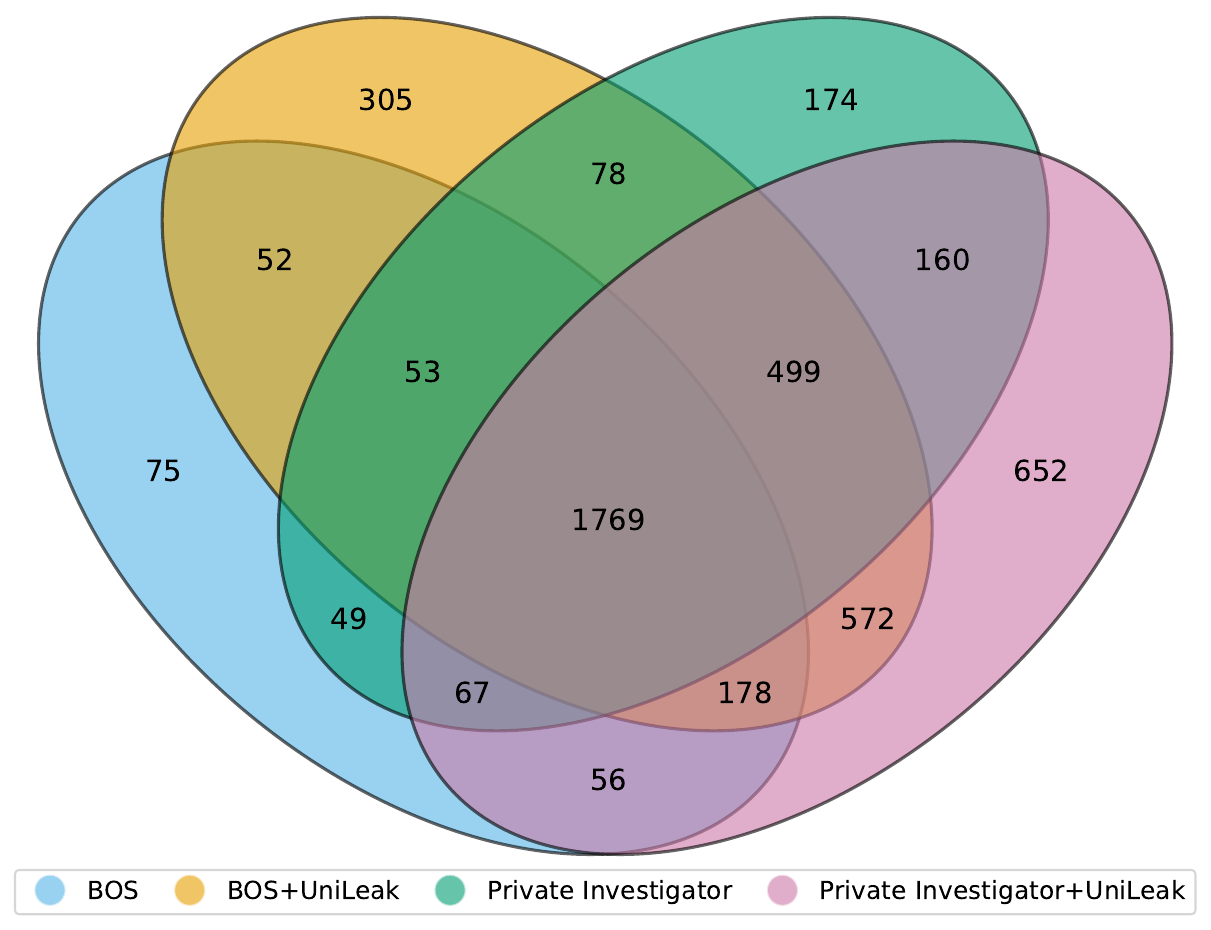}
    \end{minipage}
    \hfill
    \begin{minipage}[t]{0.32\linewidth}
        \centering
        \includegraphics[width=\linewidth]{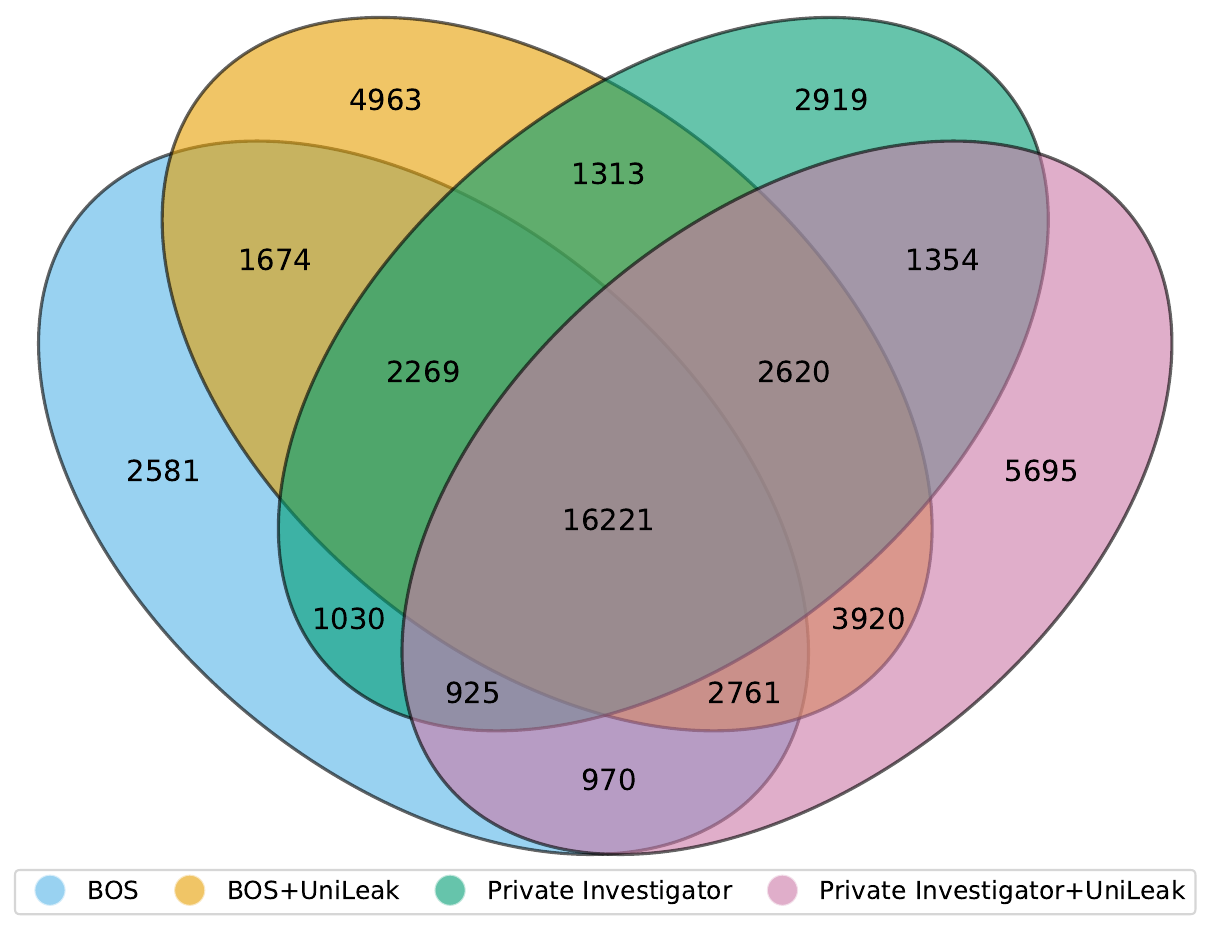}
    \end{minipage}
    \hfill
    \caption{\textbf{Unique PIIs extracted per-method on Phi-2B.}
    From Left to Right: Emails, Phones, Names. Top: TREC, Bottom: Enron}
    \label{fig:phi-phone-email-overlap}
\end{figure}

We observe that \ourmethod consistently extracts unique PII records that are inaccessible to baseline methods.
In the case of Private Investigator+\ourmethod on Phi--Enron (emails), 
the disparity is especially pronounced, 
with this combination uniquely extracting over 
1{,}457 PII items that no other method recovers.

This trend holds across multiple settings.
In 4 out of 6 Phi-2B scenarios shown in Figure~\ref{fig:phi-phone-email-overlap}, 
both BOS+\ourmethod and Private Investigator+\ourmethod 
achieve a higher number of exclusively extracted PII records 
than their respective baselines.
Notably, we do not observe a single case 
\ourmethod fails to outperform both corresponding baselines 
in terms of unique PII extraction.

\begin{figure}[ht]
    \centering
    \begin{minipage}[t]{0.32\linewidth}
        \centering
        \includegraphics[width=\linewidth]{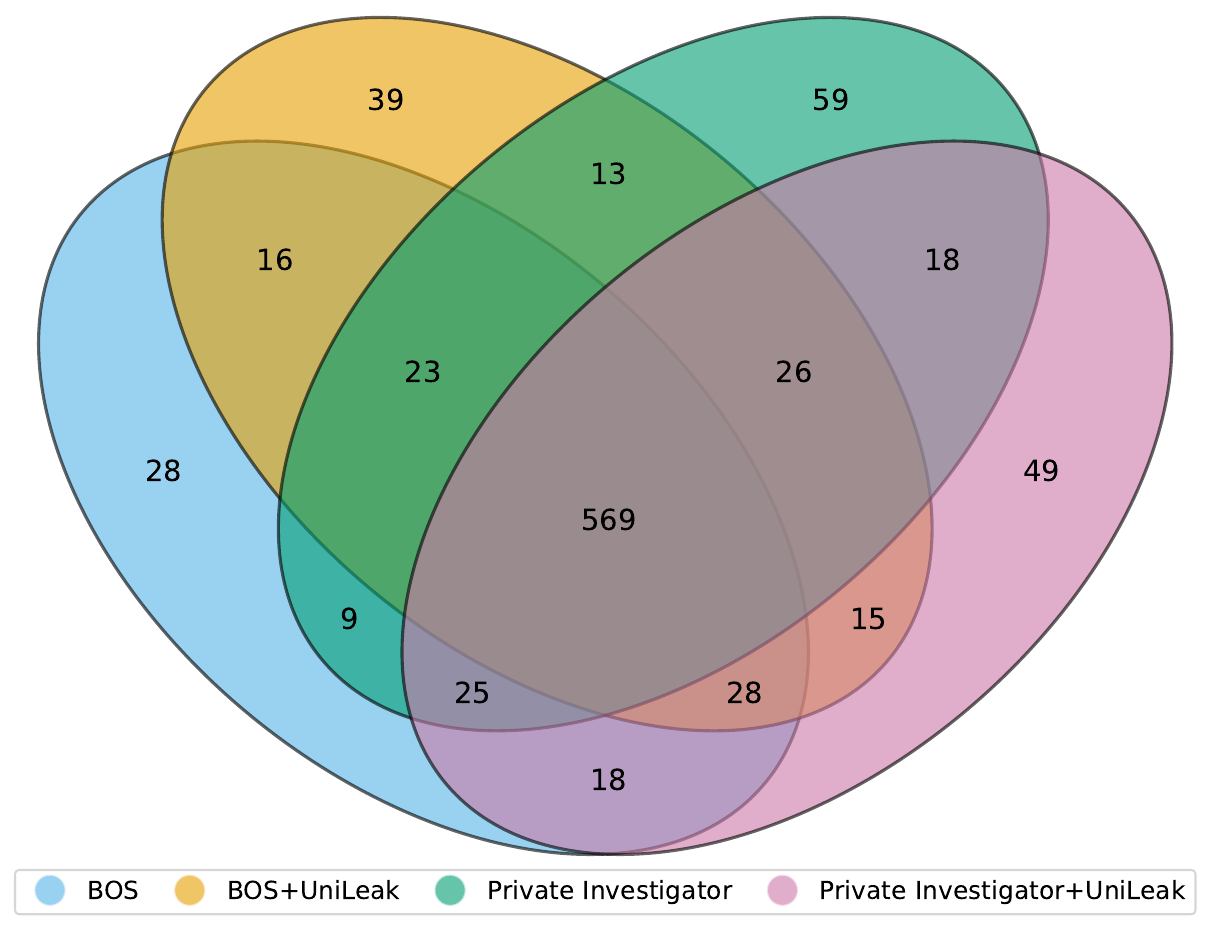}
    \end{minipage}
    \hfill
    \begin{minipage}[t]{0.32\linewidth}
        \centering
        \includegraphics[width=\linewidth]{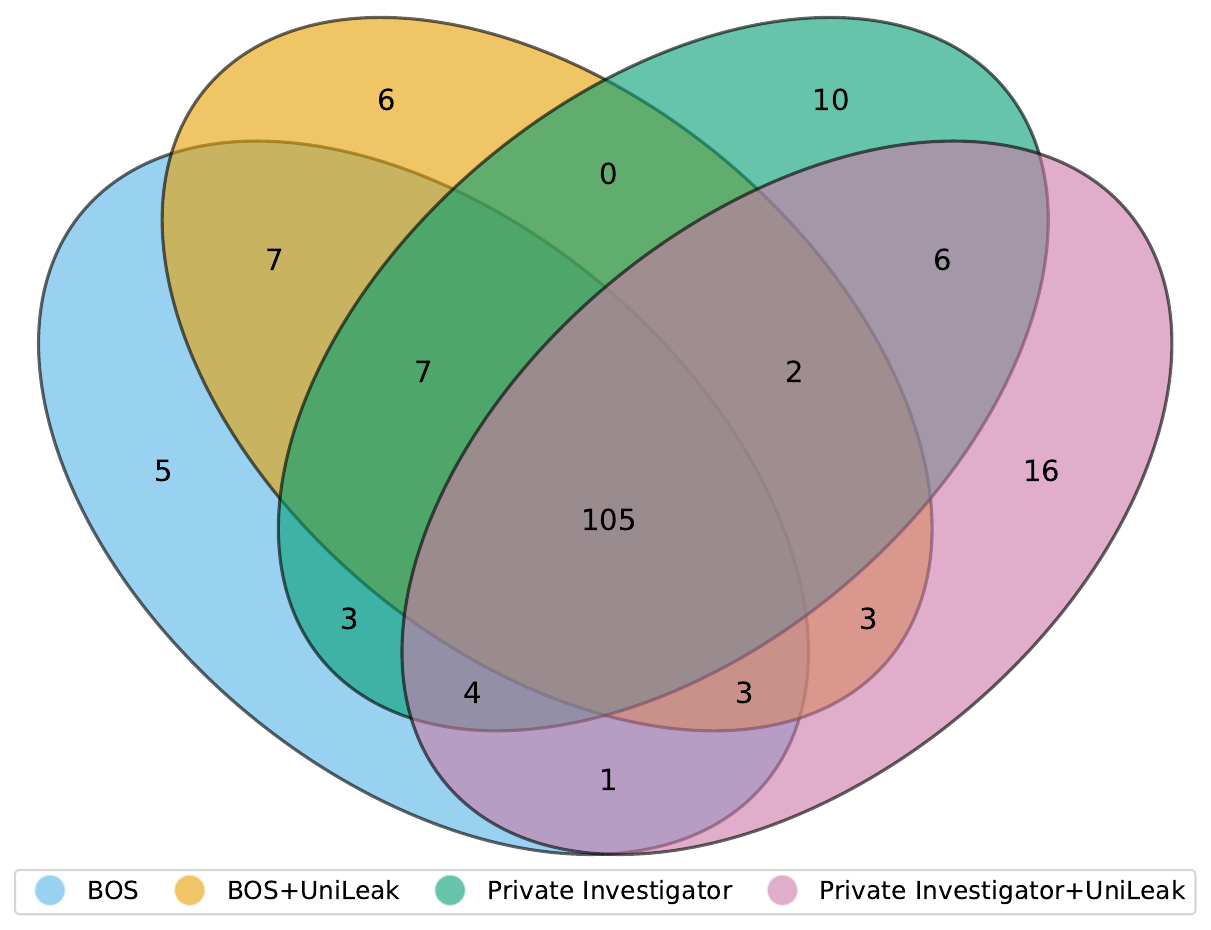}
    \end{minipage}
    \hfill
    \begin{minipage}[t]{0.32\linewidth}
        \centering
        \includegraphics[width=\linewidth]{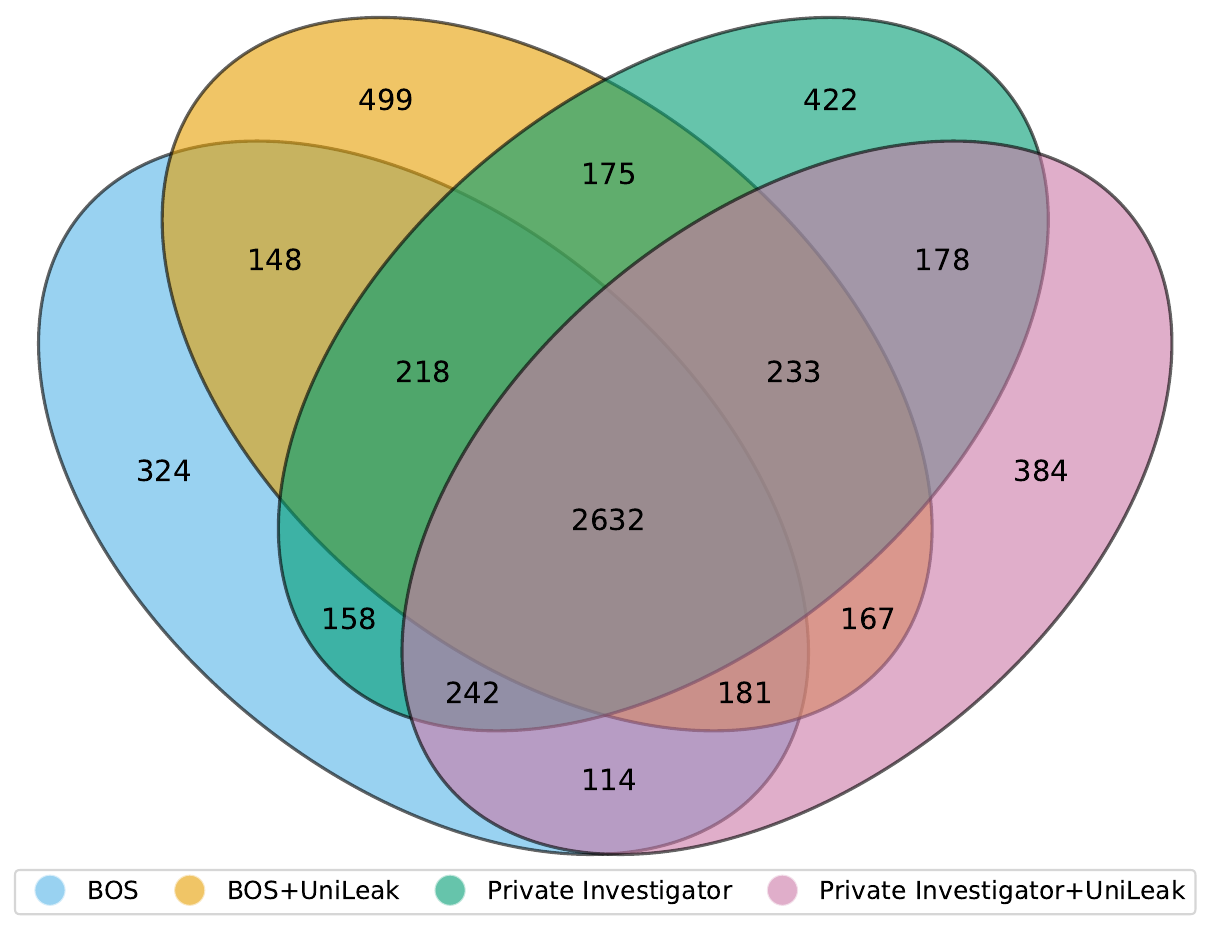}
    \end{minipage}
    \hfill
    \begin{minipage}[t]{0.32\linewidth}
        \centering
        \includegraphics[width=\linewidth]{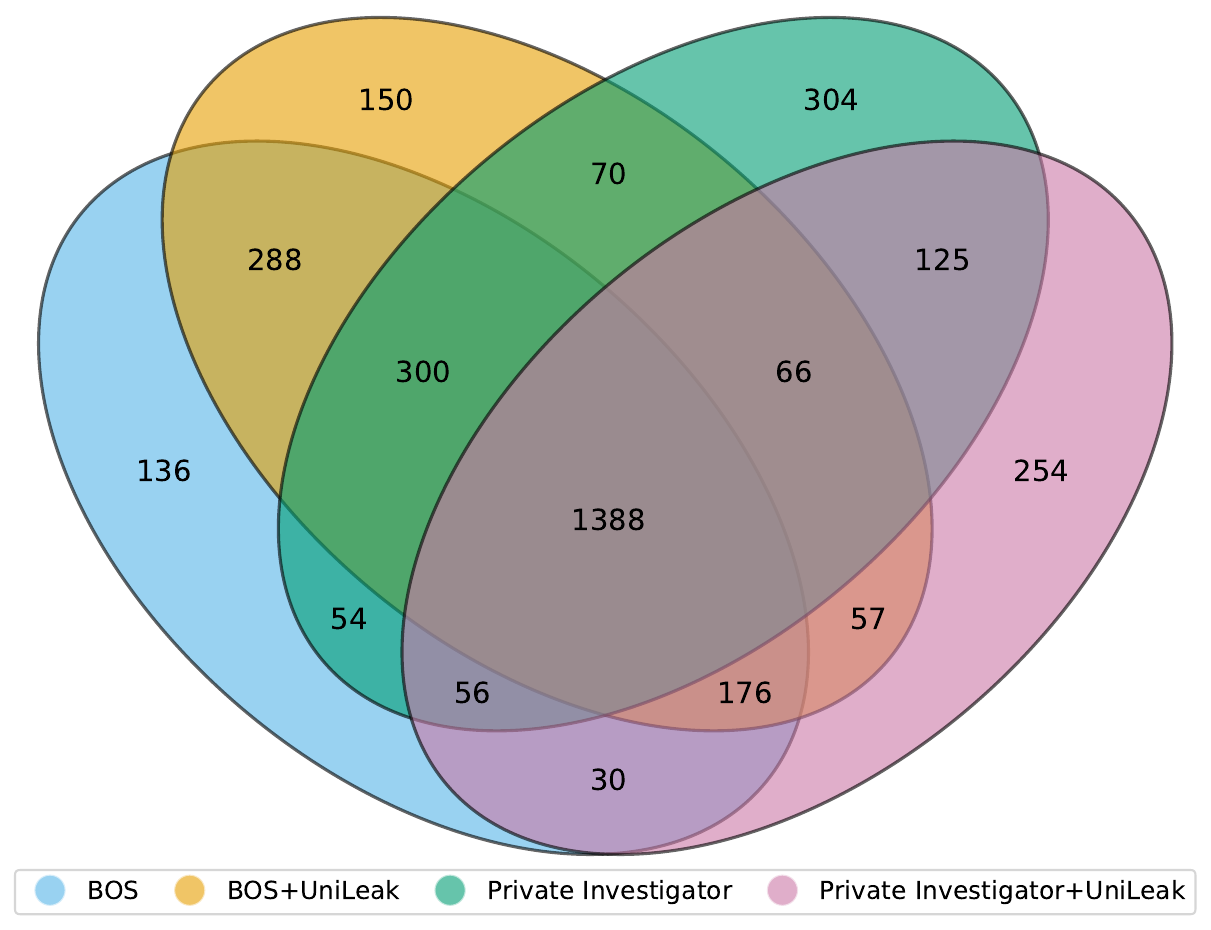}
    \end{minipage}
    \hfill
    \begin{minipage}[t]{0.32\linewidth}
        \centering
        \includegraphics[width=\linewidth]{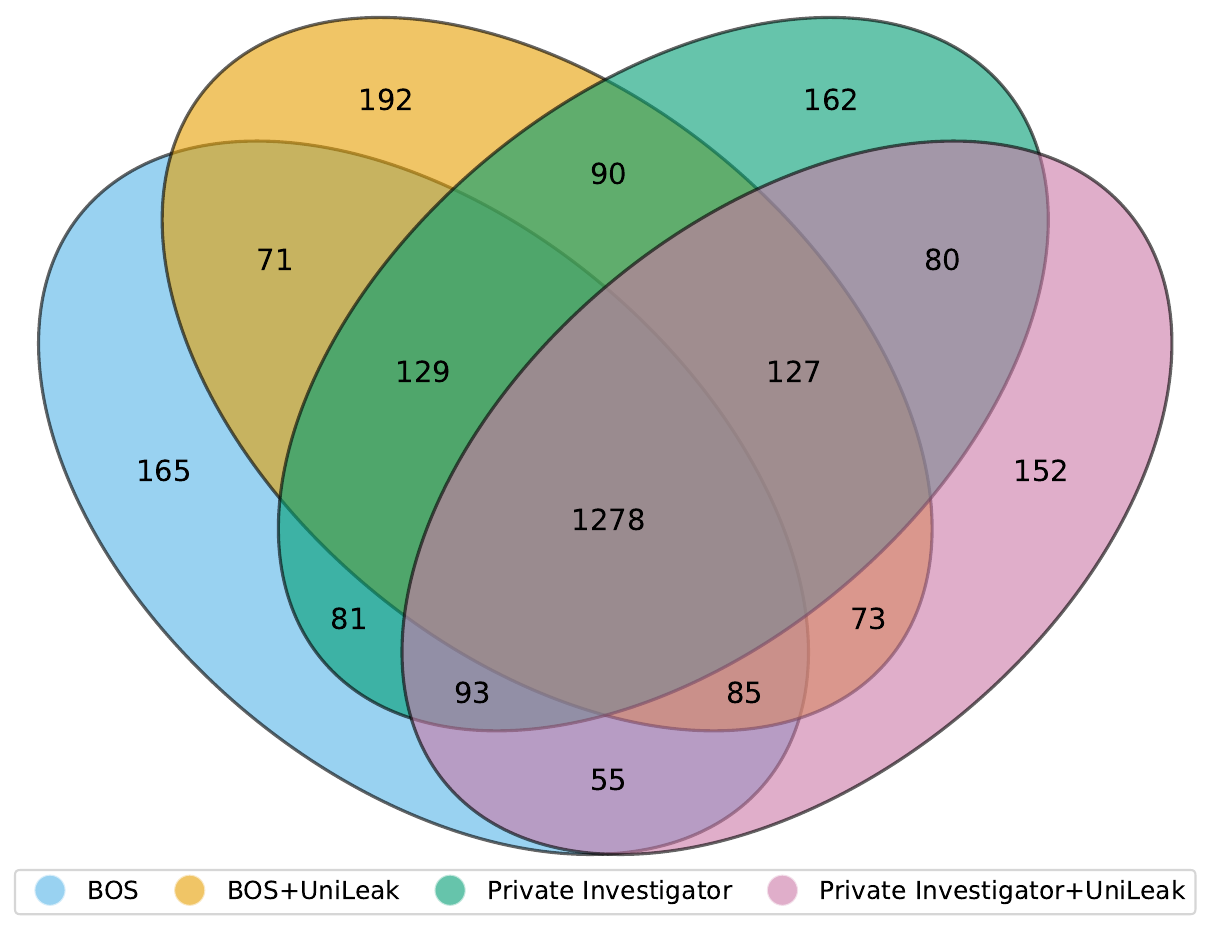}
    \end{minipage}
    \hfill
    \begin{minipage}[t]{0.32\linewidth}
        \centering
        \includegraphics[width=\linewidth]{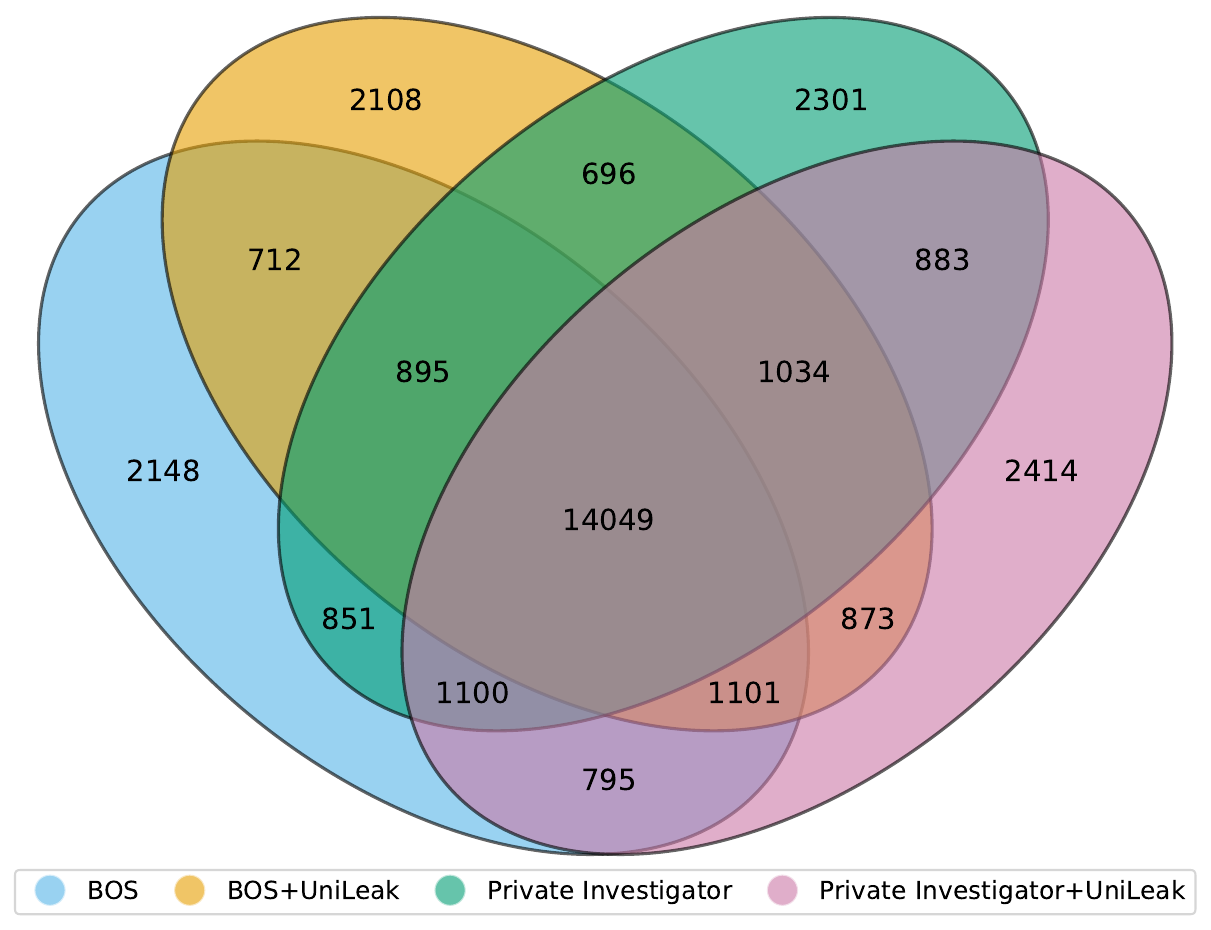}
    \end{minipage}
    \hfill
    \caption{\textbf{Unique PIIs extracted per-method on GPT-Neo}.
    From Left to Right: Emails, Phones, Names. Top: TREC, Bottom: Enron}
    \label{fig:gptneo-phone-email-overlap}
\end{figure}

In some settings, BOS+\ourmethod even exceeds 
Private Investigator+\ourmethod in extracting unique PII. 
For example, in the GPT-Neo-TREC setting for person names 
(Figure~\ref{fig:gptneo-phone-email-overlap}), 
BOS+\ourmethod recovers more unique PII items than its Private Investigator counterpart.

As expected, the largest pairwise overlap often occurs 
between a baseline method and its \ourmethod-augmented variant, 
reflecting their shared prompting mechanisms (e.g., GPT-Neo–Enron emails). 
However, we also observe multiple cases where the two \ourmethod-augmented attacks 
exhibit the highest or second-highest overlap with each other.
This suggests that \ourmethod is not merely amplifying 
prompt-specific behavior, but instead activates a model-specific representation 
that governs PII expression more broadly.

\section{Qualitative Comparison of Model Generations}
\label{appendix:qualitative-comparison}

\begin{figure}[ht]
\centering
\begin{tcolorbox}[title=BOS, colback=gray!10, colframe=black!50]
\small
-----Original Message----- \\
From: Michael R. Reagle Jr. [mailto:reagle@w3.org] \\
Sent: Wednesday, May 08, 2000 6:35 AM \\
To: w3c-query@w3.org \\
Subject: RE: The XQuery namespace spec \\

One of the advantages of XQuery-SGML over XML is that it allows \\
both a query-aware system and a reader to determine that the \\ 
namespace has been correctly formed.
\end{tcolorbox}

\begin{tcolorbox}[title=BOS + \ourmethod + First Token Steering, colback=blue!10, colframe=blue!50]
\small
\textless \textbar endoftext \textbar \textgreater \\
today I got a quick read. \\ \\ \\   
----- Original Message ----- \\

  From: "David Poehlman" \textless poehlman1@comcast.net \textgreater \\
  To: \textless www-dom@w3.org \textgreater \\
  Sent: Monday, September 12, 2001 8:20 PM \\
  Subject: Re: DOM L3 vs DOM L2
\end{tcolorbox}

\begin{tcolorbox}[title=BOS + \ourmethod + All-Token Steering, colback=blue!10, colframe=blue!50]
\small
\textless \textbar endoftext \textbar \textgreater \\                               http://www.net21.com.cn/mars/marshall/ \\
  \textbackslash \textbackslash -----------------------\\ \\ \\ 
  \textbackslash --------------------\\
  ---------------------\\ \\ 
  \textbackslash \textbackslash ======================\\ \\ \\
  -----------------------\\
  --------------------------
\end{tcolorbox}
\caption{\textbf{Qualitative comparison of generated text during extraction.} 
From top to bottom: BOS, BOS + \ourmethod{} with first-token steering (default), 
and BOS + \ourmethod{} with all-token steering.
}
\label{fig:qualitative}
\end{figure}

\topic{Steering.}
We qualitatively compare generations from the base model using BOS sampling, BOS+\ourmethod, and BOS+\ourmethod when steering is applied at all token positions rather than only the first.
Figure~\ref{fig:qualitative} presents representative examples for each setting.

When steering is applied at all token positions, output quality degrades substantially: the resulting text is incoherent and often not recognizable as any natural language.
In many cases, the only discernible structure consists of repeated tokens, with some generations degenerating into indefinite repetition.
In contrast, generations produced using BOS sampling and \ourmethod exhibit coherent structure and characteristics consistent with the training data distribution.
This further supports our decision to steer only at the first token position, as it preserves generation quality while still amplifying PII leakage.

\topic{Mitigation (subtraction)}.
We also qualitatively compare the generations obtained from our mitigation, in which we instead \emph{subtract} the directions identified by \ourmethod.
Figure~\ref{fig:qualitative-mitigation} shows representative examples.
Quality is generally preserved (first two cases), but we observe that subtracting \ourmethod's directions for Private Investigator on Enron (last case) results in degenerate text, possibly due to interactions with the optimized prompt-tokens.

\begin{figure}[ht]
\centering
\begin{tcolorbox}[title=BOS - \ourmethod, colback=gray!10, colframe=black!50]
\small
John Russell wrote:\\  I had previously asked for a simple example
   to make sure\\ the "standard" way of defining entity references is in place.\\
   In\\ this example it was suggested, but that is still a good way to go.\\
  \\ So, given a XML document I have to define an entity reference.\\ \\
    My first stab at understanding this is to have the following text\\  -\\
        - Define a "standard" way to store entities that are not on any\\  
        external storage system and only need to reference some\\       internal
   storage system.\\ \\ (This is what the "canonical" standard seems to me
   to be providing.)\\ If the entity references cannot be referenced on a "external
   storage\\ system", the "canonical" standard will fail.\\ \\ We are trying
   to solve this problem using a very simple DTD.\\ Why isn't the above an XML declaration?\\
   I cannot tell if it is a DTD or a schema file, because I could not\\
\end{tcolorbox}

\begin{tcolorbox}[title=Private Investigator - \ourmethod on TREC, colback=blue!10, colframe=blue!50]
\small
OnChange is an HTML Tidy compliant XHTML\\ XFORM which is used within Tidy (vers 4th October version,\\ April 2000-May, 2000, available at:\\ http://www.w3.org/TR/xhtml1/\#validate)\\\\If any of the text is missing, Tidy will attempt this on the entire \\ DOM with a complete text file with all required attributes added \\for the purpose of determining the validity of the content. For example,\\as below, a text file with all attributes missing is added but the \\validation of this data fails. Tidy will check the contents of the \\textfile with the correct attributes and the code will fail and attempt\\to write the text for editing.\\\\\\This is done for convenience only as a way to limit the use of \\"non-HTML" tag. We believe that this option will aid in reducing \\trouble with validating Tidy's.\\\\Regards, Terry Teague\\\\--\\\*****************************************************\\Terry Teague                     Phone: 800-258-8857
\end{tcolorbox}

\begin{tcolorbox}[title=Private Investigator - \ourmethod on Enron, colback=blue!10, colframe=blue!50]
\small
\t)' retali installmentsments pay R R R R R R R R R R R R R R R R R R R R R R R R R R R R R R R R R R R R R R R R R R R R R R R R R R R R R R R R R R R R R R R R R R R R R R to R R R R R R R R R R R R R R R R R R R R R R R R R R R R to R R R R R R R R R R R R R R R R R R R R R R R R R R R R R R R R R R R R R R R R R R R R R R R R R R R R R R R R R R R R R R R R R R R RR R R R R R R R R R R R R R R R R R R R R R R R R R R R R R R R R R R R R R R R R R R R R R R R R R R R R R R R R R R R R R R R R R R R R R R R R R R R R R R R R R"
\end{tcolorbox}
\caption{\textbf{Qualitative comparison of generated text during mitigation} on GPT-Neo. From top to bottom: BOS + \ourmethod on TREC, Private Investigator + \ourmethod on TREC, and Private Investigator + \ourmethod on Enron.
}
\label{fig:qualitative-mitigation}
\end{figure}

\end{document}